# Selection of contributing factors for predicting landslide susceptibility using machine learning and deep learning models

Cheng Chen[a], Lei Fan[a,*]

[a] *Department of Civil Engineering, Xi'an Jiaotong-Liverpool University, Suzhou 215123, China*

[*] *Correspondence: lei.fan@xjtlu.edu.cn*

**Abstract**

Landslides are a common natural disaster that can cause casualties, property safety threats and economic losses. Therefore, it is important to understand or predict the probability of landslide occurrence at potentially risky sites. A commonly used means is to carry out a landslide susceptibility assessment based on a landslide inventory and a set of landslide contributing factors. This can be readily achieved using machine learning (ML) models such as logistic regression (LR), support vector machine (SVM), random forest (RF), extreme gradient boosting (Xgboost), or deep learning (DL) models such as convolutional neural network (CNN) and long short time memory (LSTM). As the input data for these models, landslide contributing factors have varying influences on landslide occurrence. Therefore, it is logically feasible to select more important contributing factors and eliminate less relevant ones, with the aim of increasing the prediction accuracy of these models. However, selecting more important factors is still a challenging task and there is no generally accepted method. Furthermore, the effects of factor selection using various methods on the prediction accuracy of ML and DL models are unclear. In this study, the impact of the selection of contributing factors on the accuracy of landslide susceptibility predictions using ML and DL models was investigated. Four methods for selecting contributing factors were considered for all the aforementioned ML and DL models, which included Information Gain Ratio (IGR), Recursive Feature Elimination (RFE), Particle Swarm Optimization (PSO), Least Absolute Shrinkage and Selection Operators (LASSO) and Harris Hawk Optimization (HHO). In addition, autoencoder-based factor selection methods for DL models were also investigated. To assess their performances, an exhaustive approach was adopted, testing all possible selection cases of contributing factors, the results of which served as the benchmark. The results confirmed that using more important contributing factors improved the prediction accuracy of the ML models considered. However, it was interesting to find that the selection of contributing factors using IGR and RFE reduced the predictive power of the DL models. For the DL models, using an autoencoder architecture improved their prediction performance. The study also found that the choice of factor selection methods was far more effective than the selection of contributing factors in improving the prediction accuracy of landslide susceptibility.

*Keywords:* landslide; prediction; landslide susceptibility; contributing factors; factor selection; machine learning; deep learning;

## 1. Introduction

Landslides are a common natural hazard, in which large volumes of soil and/or rocks are displaced downward (Leynaud, Mulder et al. 2017). It can lead to varying degrees of infrastructure damages and/or personal injuries (Peduto, Santoro et al. 2021), in addition to various negative impacts on social and environmental sustainability. Landslide prediction is an important task for landslide prevention and control, and it is the basis for deciding whether to take measures to alleviate the possible adverse effects of landslides. To this end, landslide susceptibility mapping (LSM) is often considered to predict the spatial likelihood of landslide occurrence (Reichenbach, Rossi et al. 2018), where the classification, size and spatial distribution of potential landslides are assessed. It is also a useful and proactive means of delineating landslide risky areas.

To produce LSM, a wide range of qualitative or quantitative models have been proposed. Qualitative models rely solely on expert judgment and involve direct mapping of the landscape to assess susceptibility against expert-defined factors (Aleotti and Chowdhury 1999, Tanyu, Abbaspour et al. 2021). Quantitative models mainly refer to statistical, machine learning (ML) or deep learning (DL) models, such as support vector machine (SVM) (Marjanović, Kovačević et al. 2011), logistic regression (LR) (Sun, Xu et al. 2021), random forests (RF) (Kim, Lee et al. 2018), extreme gradient boosting (Xgboost) (Can, Kocaman et al. 2021), convolutional neural network (CNN) (Wang, Fang et al. 2019, Sameen, Pradhan et al. 2020, Zhao, Chen et al. 2021), and long short time memory (LSTM) (Fang, Wang et al. 2021).

Contributing factors to landslides are diverse and typically include topographical, hydrological, geological, land-cover and hydrological contributing factors (Hong, Adler et al. 2007). Although existing data-driven models have shown their ability to achieve good prediction accuracy (Gaidzik and Ramírez-Herrera 2021), taking too many factors into account in those models may not only lead to overfitting but also affect model generalization (Kotsiantis 2011, Vasu and Lee 2016). In addition, it may require more computing time. Therefore, selecting some factors that are more relevant to a model is a rational approach to consider. To this end, factor selection methods are needed. Some limited studies have shown the effectiveness of factor selection for landslide susceptibility modeling (Micheletti, Foresti et al. 2014).

For ML models, factor selection methods can be categorized into three types: filter, wrapper and embedded methods. Filter methods (Malekipirbazari, Aksakalli et al. 2021) work independently of a particular classifier. It measures the specific attributes of data, such as relevance, distance or information gain, and ranks the data based on those attributes to eliminate less relevant or irrelevant ones. Acharya (Acharya 2018) used chi-square (CS), correlation, information gain (IG) and information gain ratio (IGR) to select the factors that were more suitable for landslide susceptibility assessment at the regional scale in Nepal. Pham et al. (Pham, Van Dao et al. 2021) used CS to select some more appropriate input factors to train an ANN model for landslide susceptibility assessment and mapping. Filter methods are interpretable and computationally fast, however in such methods the selection of factors and their degrees of importance are tied to a specific supervised learning algorithm. Wrapper-based factor selection methods generate subsets of factors, based on searching strategies combined with learning algorithms as forward or backward selection, or heuristic selection of factors. In embedded methods, factor selection is an embedded part of the learning training process, making it easier to obtain an optimal subset of factors in the learning process (Song, Wang et al. 2020).

Although factor selection methods had been used to improve the prediction accuracy of ML models (Kotsiantis 2011, Micheletti, Foresti et al. 2014, Vasu and Lee 2016, Malekipirbazari, Aksakalli et al. 2021, Pham, Van Dao et al. 2021, Tanyu, Abbaspour et al. 2021), there is still a lack of research specifically comparing the impact of various selection methods on ML models. For DL models, some researchers also considered factor selection. For example, Wang et al. (Wang, Fang et al. 2019) used the gain ratio for factor selection in a 1D-CNN model for LSM. Hakim et al. (Hakim, Rezaie et al. 2022) applied downsampling through the pooling process to reduce the dimensionality of factor classes to achieve factor selection. Li et al. (Li and Becker 2021) used an autoencoder for factor extraction to avoid removing useful information. However, it is unknown whether factor selection will improve the prediction accuracy of DL models. As such, it would be useful to investigate the effect of factor selection on the prediction accuracy of ML and DL models for LSM.

This study aims to comprehensively investigate the impact of the selection of contributing factors on the accuracies of mainstream ML and DL models for predicting landslide susceptibility. Specifically, we demonstrate the impact of various factor selection methods, such as IGR (Dou, Yunus et al. 2020), Particle Swarm Optimization (PSO) (Wang, Wang et al. 2023), Recursive Feature Elimination (RFE) (Zhou, Wen et al. 2021), Least Absolute Shrinkage and Selection Operators (LASSO) (Ghosh, Azam et al. 2021), and Harris Hawk Optimization (HHO) (Long, Jiao et al. 2022) on the prediction accuracies of some typical ML and DL classification models, including LR, SVM, RF, Xgboost, CNN, and LSTM. We also compare the effectiveness of these factor selection methods with a novel autoencoder-based factor selection method specifically designed for DL models. Furthermore, an exhaustive factor selection approach tests all possible cases of contributing factor selection as a benchmark for assessing the performance of the factor selection methods. Our findings provide practical guidance for practitioners seeking to select an appropriate factor selection method to enhance the accuracy of landslide susceptibility assessment using ML and DL models. Overall, this research contributes to the field by advancing our understanding of the selection of contributing factors in landslide prediction and provides a comprehensive approach to improving the accuracy of landslide susceptibility assessments.

This article is structured as follows. Section 2 presents the study data and the landslide contributing factors considered in this study. Section 3 mainly details the factor selection methods and ML and DL models used for LSM. Section 4 introduces the implementation of the experiments, mainly including the evaluation criteria, the setting of model hyperparameters, and the parameters of the factor selection methods. Detailed experimental results and a comparison of the LSM outputs are reported in Section 5 and Section 6, respectively. Some discussions are carried out in Section 7, followed by conclusions in Section 8.

## 2. Study site and data

### 2.1. Study area

Kerala is the chosen study area to evaluate the factor selection methods. It is located in the southwestern part of the Indian Peninsula, on the windward slope of the Western Ghats, on the eastern coast of the Arabian Sea (Hao, Rajaneesh et al. 2020), as shown in Fig.1. This region has a typical tropical climate with a monsoon season. The bedrock in Kerala is highly weathered, resulting in thick, poorly consolidated soils (mainly clay) covering most of the region (Sajinkumar, Anbazhagan et al. 2011). In terms of geomorphology, it has rugged mountains with deep valleys and plateaus in the east and plains along the western coast (Vishnu, Sajinkumar et al. 2019). The Western Ghats are dominated by ancient fault cliffs on the plateau, often with very steep slopes that are prone to slope instability (Kuriakose, Sankar et al. 2009). The study area map (shown in Fig.1) was created using Esri's ArcMap software. For clearer visual inspection, two local regions (i.e., Region 1 and Region 2 shown in Fig. 1) were selected and zoomed in to examine the risk categories in the landslide susceptibility map against recorded historical landslides.

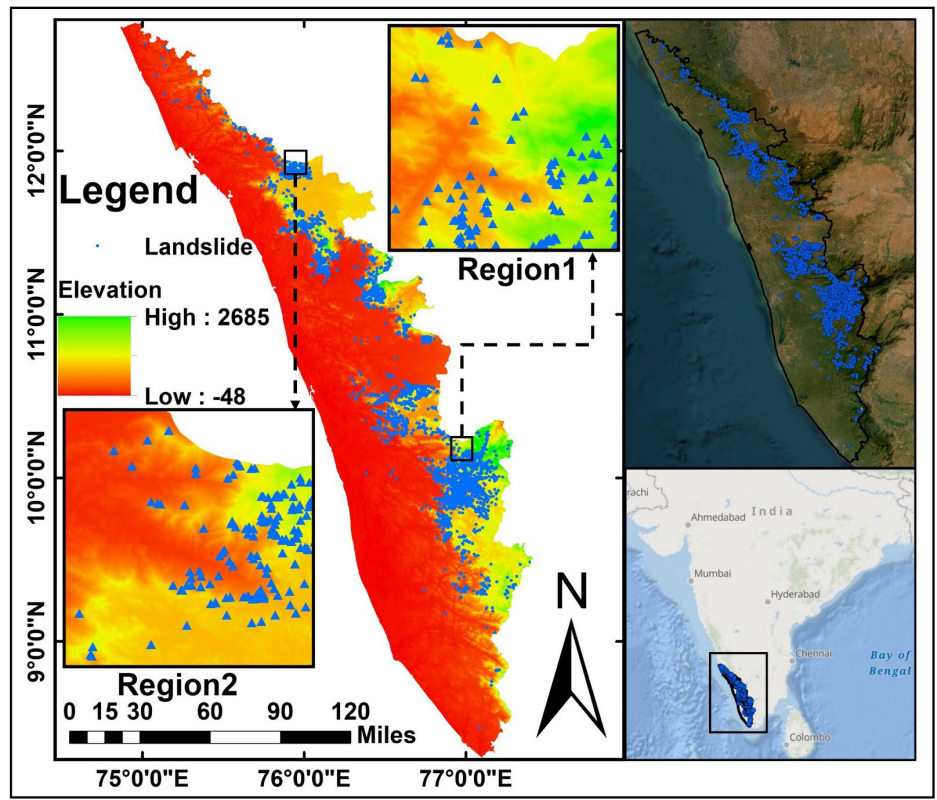

Fig. 1. Information on the study area Kerala, including its location, the digital elevation map (i.e., the heatmap), and recorded historical landslide cases (represented by blue dots or triangles)

### 2.2. Study data and landslide contributing factors

Kerala's historical landslide inventory was established by the National Remote Sensing Centre (NRSC) of the Indian Space Research Organization and Geological Survey of India. It is available from NRSC' website. As of the time the dataset was cited in this study, the dataset recorded a total of 4728 identified landslides, as shown in Table 1. These landslide points are illustrated by scattered blue markers, as shown in Fig. 1. Remote sensing images of the study area were obtained from Google Earth Engine (Gorelick, Hancher et al. 2017) and the Indian Institute of Remote Sensing (Ramasamy, Gunasekaran et al. 2021). The DEM map with a spatial resolution of 30 m was obtained from NASA SRTM Digital Elevation.

Table 1. Number of landslides by type and percentage of all landslides occurring in Kerala during monsoon 2018 (Hao,

Rajaneesh et al. 2020)

| District | Debris flows | | Shallow slides | | Rock fall | | Total number |
|---|---|---|---|---|---|---|---|
| | Number | Percent | Number | Percent | Number | Percent | |
| Idukki | 679 | 31% | 1421 | 64% | 123 | 5% | 2223 |
| Palakkad | 699 | 88% | 92 | 12% | 2 | 0% | 793 |
| Malappuram | 358 | 84% | 66 | 15% | 5 | 1% | 429 |
| Wayanad | 252 | 76% | 68 | 21% | 11 | 3% | 331 |
| Thrissur | 234 | 91% | 20 | 8% | 2 | 1% | 256 |
| Kozhikode | 204 | 89% | 18 | 8% | 6 | 3% | 228 |
| Kannur | 120 | 85 | 19 | 14 | 2 | 1 | 141 |
| Ernakulam | 96 | 90 | 11 | 10 | 0 | 0 | 107 |
| Pathanamthitta | 92 | 87 | 13 | 12 | 1 | 1 | 106 |
| Kottayam | 65 | 86 | 11 | 14 | 0 | 0 | 76 |
| Kasaragod | 5 | 21 | 19 | 79 | 0 | 0 | 24 |
| Kollam | 9 | 90 | 1 | 10 | 0 | 0 | 10 |
| Thiruvananthapuram | 3 | 75 | 1 | 25 | 0 | 0 | 4 |
| **Total in Kerala** | **2816** | **60** | **1760** | **37** | **152** | **3** | **4728** |

Selecting suitable contributing factors is a key step in LSM (Sameen, Pradhan et al. 2020). In this study, 15 contributing factors were considered, including topography factors (i.e. elevation, aspect, slope, plan curvature and profile curvature), geological factors (i.e. lithology, distance to faults), land-cover factors (i.e. land use, distance to roads, normalized difference vegetation index (*NDVI*)), and hydrological factors (i.e. distance to stream, rainfall (Lee and Talib 2005), sediment transport index (*STI*), stream power index (*SPI*), topographic wetness index (*TWI*)).

Topography factors are derived using a DEM. Elevation affects the degree of rock weathering and is considered an important factor in the landslide susceptibility analysis (Yao, Tham et al. 2008). Slope influences the formation of landslides, which regulates moisture concentrations on terrain surfaces (Sameen, Pradhan et al. 2020). Curvature represents the change in slope of a topographic surface along the small arc of the curve (Zhao, Masoumi et al. 2022), and can affect the slope stability (Ohlmacher 2007). Profile curvatures and plan curvatures are the curvatures in the slope direction and contour direction, respectively. They also affect the rate of descent of water and the degree of weathering of slope rock.

Lithology affects landslide occurrences because different types of rock have different resistances to weathering, and certain types of rock are more prone to landslides. Some landslides occur along existing faults where tectonic movements have reduced rock strength and created permeable fracture zones, which increases the likelihood of landslide occurrences.

In this study, land use is classified as agricultural land, woodland, grassland, water or residential, representing different levels of regional development that may directly or indirectly affect landslide occurrence (Mugagga, Kakembo et al. 2012). For slopes close to roads, the road construction itself and the subsequent surcharge loading carried by the road can lead to ground instability. *NDVI* is considered because plant roots not only store or release water during evaporation, which contributes significantly to slope hydrology, but also act as reinforcements for slope materials to strengthen slope stability (Fiorucci, Ardizzone et al. 2019).

Hydrological factors are known to be important contributors to landslide occurrence (Bordoni, Meisina et al. 2015). Rainfall is often a trigger for many landslides (Rong, Li et al. 2020), and the maximum precipitation in August 2018 is used in this study. The distance from streams may also affect landslide occurrences. The shorter the distance to a stream, the more likely the slope stability will be affected, as water flow may erode the slope (Lin, Chang et al. 2017). Sediment Transport Index (*STI*) reflects the impact of river erosion on slopes (Sun, Wen et al. 2021). A larger *STI* value indicates that most of the slope deposits are loose. Meanwhile, it is also prone to slope damage and deposition. Stream Power Index (*SPI*) is described as the motion of strong particles when gravity acts on sediments. Topographic Wetness Index (*TWI*) measures the topographic control of hydrologic processes, reflecting slope and flow directions. The equations of *SPI*, *STI* and *TWI* are shown in Eq. 1, Eq. 2 and Eq. 3, respectively (Zhao and Chen 2020).

$$SPI = \alpha \tan \beta \quad (1)$$

$$STI = \left(\frac{A_s}{22.13}\right)^{0.6} \times \left(\frac{\sin\beta}{0.0896}\right)^{1.3} \quad (2)$$

$$TWI = ln\left(\frac{A_s}{\beta}\right) \quad (3)$$

where $A_s$ is the specific catchment area; $\beta$ is the slope angle in degree; $\alpha$ is the specific catchment area ($A = A/L$, catchment area ($A$) divided by contour length ($L$)).

From the study dataset presented in the first paragraph of Section 2.2, the maps of the aforementioned landslide contributing factors can readily be derived. The maps of those factors are shown in Fig. 2, which are essentially the input data to LSM. The statistical graphs of the contributing factors are shown in Fig. 3.

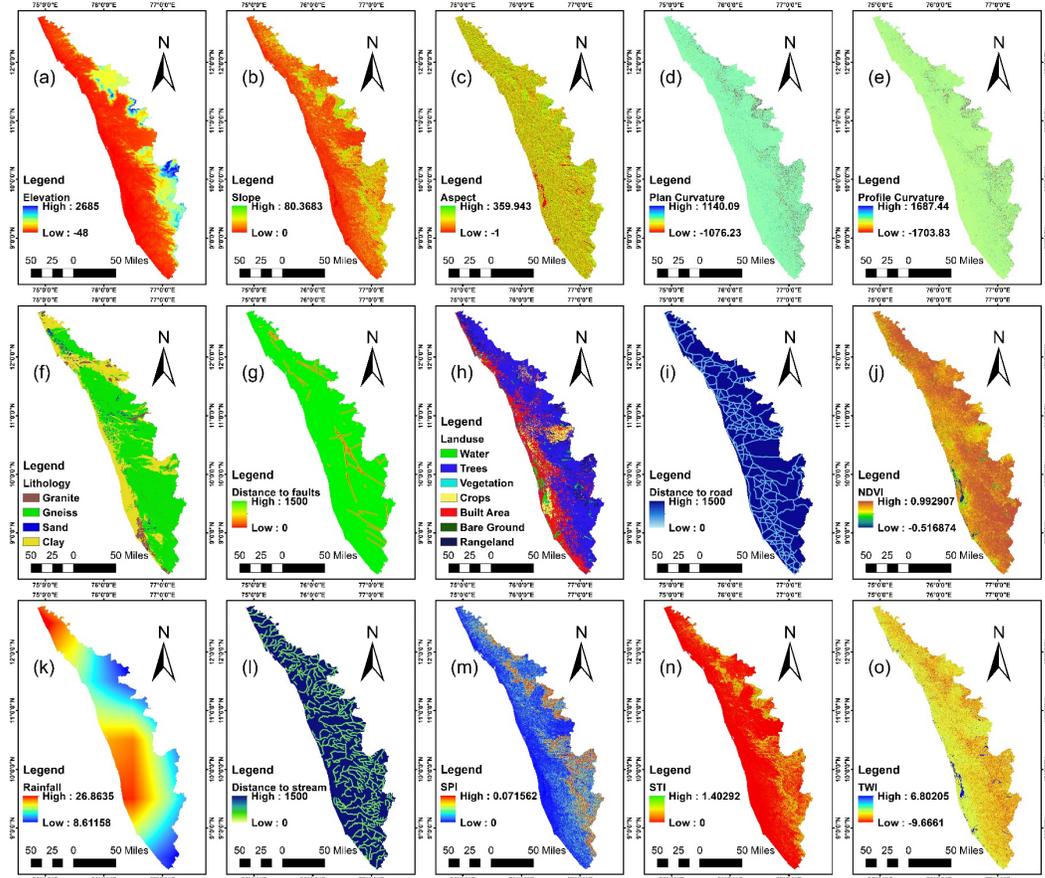

Fig. 2. Maps of the contributing factors considered: (a) elevation, (b) slope, (c) aspect, (d) plan curvature, (e) profile curvature (f) lithology, (g) distance to faults, (h) land use, (i) distance to road, (j) *NDVI*, (k) rainfall, (l) distance to stream, (m) *SPI*, (n) *STI* and (o) *TWI*.

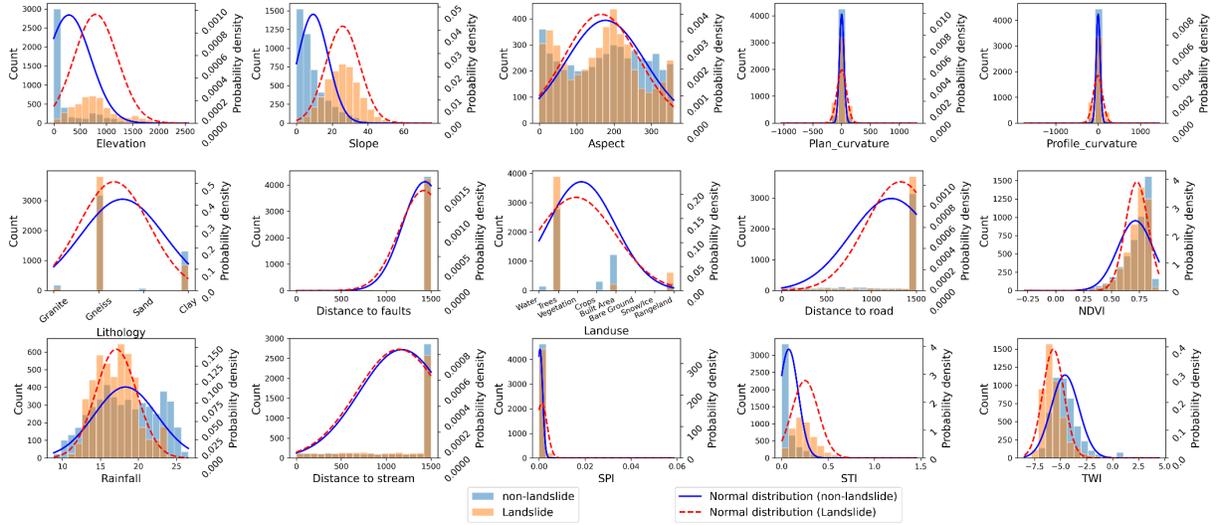

Fig. 3. Statistical distributions of individual contributing factors

## 3. Methodology

### 3.1. Classification model

In our study, we considered six different mainstream classification models and used the optimized parameter settings of those classifiers in their respective implementations in the Scikit-Learn module or TensorFlow module. These classifiers include SVM, LR, RF, Xgboost, LSTM and CNN. Their fundamental principles are briefly described as follows:

- SVM: It is a supervised learning model for mapping instances to separate different classes with hyperplanes. It constructs hyperplanes in a high-dimensional space and classifies new instances according to the side of the hyperplane.
- LR: It is a supervised learning model that first fits a decision boundary (not limited to linear but also polynomial) and subsequently establishes a probabilistic link between the decision boundary and the classification, which yields the probability in the dichotomous case.
- RF: It is a bagging-based integrated learning model that constructs several random decision trees. It constructs multiple decision trees and uncorrelates the decision trees by randomly splitting the attributes of each tree. This reduces overfitting by averaging the results of mutually dissimilar trees, while maintaining the predictive power of the tree.
- Xgboost: It is an integrated boosting-based learning model that can be considered an iterative functional gradient descent algorithm. Similar to other boosting methods, Xgboost combines weak learners into a single strong learner in an iterative manner. However, Xgboost adds a regularization term to penalize the complexity of a model for ensuring good training effects and less overfitting.
- LSTM: LSTM networks are a special kind of RNN that is capable of learning long-term dependencies. Unlike simple RNNs, LSTM networks have built-in mechanisms to control how information is remembered or discarded throughout the learning process.
- CNN: It is a deep learning model of a multilayer perceptron and is commonly used for image recognition. CNN can process multiple feature maps as input and extract factors for classification through convolutional layers, Rectified Liner Units (ReLU) activation layers, and fully connected layers.

### 3.2. The architecture of the factor selection

Four architectures are usually considered for factor selection (as shown in Fig. 4). In the literature (Kotsiantis 2011, Romero, Gatta et al. 2015, Li and Becker 2021), filters, wrappers and embedded methods are commonly used to select contributing factors for ML models. In contrast, autoencoder is commonly used for DL models. The architecture of the filter method is shown in Fig. 4 (a), which operates independently of the subsequent classification model. It determines more important factors by calculating the correlations between data. Fig. 4 (b) shows the work process of the wrapper method. Factors are selected by calculating the best target score (normally, prediction accuracy) by a limited number of iterations using an ML algorithm. Usually, the optimized factor searching process stops when the results start to get worse or when the results reach a predefined threshold. Different factors may be selected by different classifiers and different predefined termination conditions. As illustrated in Fig. 4 (c), the embedded method represents a selection strategy for changing the built-in variables of the model. In this strategy, factors are neither selected nor excised. Specific controls are applied to the parameter values (i.e. weights) of the model. One weight control technique is LASSO where regularization is performed and certain weight coefficients converge to zero. When the coefficients fall to zero, they are discarded/rejected. The architecture shown in Fig. 4 (d) is referred to as the autoencoder method. It is typically used in DL models where the input data are converted to a compressed representation, but does not specifically show which factors are selected.

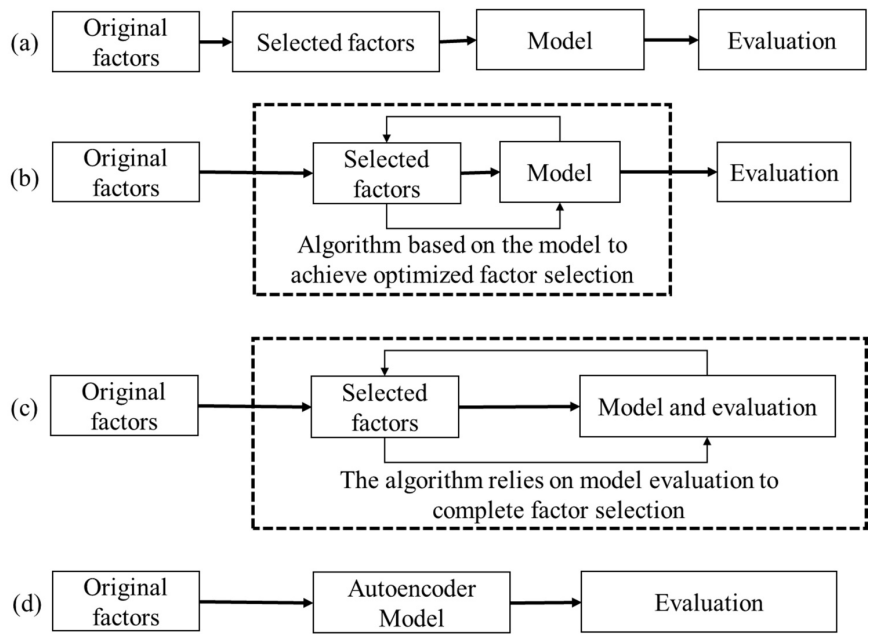

Fig. 4. Flow charts of the four architectures for selecting contributing factors: (a) filters (b) wrappers (c) embedded methods (d) autoencoders

*3.3. Factor selection methods*

All the factor selection methods investigated in our study are shown in Table 2. Some technical details of each method are provided in this section.

- **The exhaustive method**
All possible selection cases (i.e., 32767 cases in total) of the 15 contributing factors specified in Section 2.2 are enumerated and used as the inputs to the ML models considered in this study. Its purpose is to understand the performances of all possible selection cases, including the optimal one. These will be used as the benchmark to assess those achieved by the factor selection methods. Due to the computation cost of DL for each selection case and a large number of selection cases, the exhaustive method is not applied to the DL models in this study.
- **IGR**
IGR is a filter factor selection method, which is defined as how much information a contributing factor can bring to the classification system. A factor with a higher IGR indicates a higher prediction power for the model. Assuming that the training set S contains $n$ classes, the expected information is calculated using Eq. 4 where $p_i$ is the probability that a sample belongs to the class $C_i$. Factor $A$ has $m$ values, and its average entropy is calculated using Eq. 5. The split information value represents the potential information obtained by dividing $S$ into $m$ parts corresponding to the m outcomes on attribute $A$ (Eq. 6). Finally, the variable importance (*VI*) is defined in Eq. 7.

$$H(S) = -\sum_{i=1}^{n} p_i \log_2(p_i) \tag{4}$$
$$E(A) = -\sum_{i=1}^{m} p_i H(S) \tag{5}$$
$$SplitInfo_A(S) = -\sum_{i=1}^{m} X_i(S_i/S) \log_2(S_i/S) \tag{6}$$
$$VI(A) = \frac{H(S) - E(A)}{SplitInfo_A(S)} \tag{7}$$

- **RFE**

RFE is a wrapper factor selection method, which is combined with a specific classifier (typically nonlinear SVM), and has been found to be effective in factor selection (Richhariya, Tanveer et al. 2020). The RFE starts the selection using all factors available, followed by the backward elimination method that removes factors iteratively until a stopping condition of the selection criterion is reached. In every iteration of RFE, the contributing factor with the smallest weight is identified and eliminated. The essence of RFE is to train the classifier using the entire set of contributing factors, and to identify those that result in minimal decreasing margins.

- **PSO**

PSO is a hybrid method based on wrappers. PSO is an optimization algorithm that provides a subset of factors as the input to a classification model to detect the best factor selection. The basic idea of PSO is that a population of particles (candidates of factor selection) moves through a factor search space. Each particle is represented by a binary vector of length $N$ to present whether to select the corresponding factor, where $N$ denotes the total number of original factors. The initialization of factor selection of each particle is random. The motion of each particle is guided by its own known optimal position and the known optimal position of the whole population. PSO performs the optimal search by iteratively updating the velocity of the particles in the population. Each searching particle has its current velocity in the PSO, its own past best solution $\vec{p}(t)$ and the current global best solution $\vec{g}(t)$. Based on such information, the velocity of each particle is updated so that it is closer to both the global best and the past best solutions. The velocity update is performed according to Eq. 8.

$$\vec{v}(t+1) = \omega \vec{v}(t) + c_1 r_1 \left(\vec{p}(t) - \vec{x}(t)\right) + c_2 r_2 \left(\vec{g}(t) - \vec{x}(t)\right) \tag{8}$$

where $c_1$ and $c_2$ are predefined constants that determine $\vec{p}(t)$ and $\vec{g}(t)$. $\vec{v}(t)$ is the velocity of the particle and $\vec{x}(t)$ is the current particle position. $r_1$ and $r_2$ are random numbers in the interval [0, 1], and $\omega$ is a constant ($0 \leq \omega \leq 1$).

The new position is calculated by adding the previous position and the new velocity as follows. This iterative process is repeated until the stopping condition is satisfied to Eq.9.

$$\vec{x}(t+1) = \vec{x}(t) + \vec{v}(t+1) \tag{9}$$

- **LASSO**

LASSO aims to improve the predictive accuracy of a regression model by adding a penalty $\lambda \sum_{j=1}^{n} |\beta_j|$ to its loss function. The loss function is presented in Eq. 10, where $y$ is a vector of dependent variables; $x$ denotes the independent variable; $\beta$ is the corresponding coefficient. Its advantage is that it reduces some of the less critical factor coefficients to zero, based on which the less relevant factors can be removed.

LASSO is essentially a linear model that estimates sparse coefficients, and it is often used in combination with linear classifiers for factor selection. As such, it is considered for only LR and SVM (linear kernel) in Table 1.

$$loss = \sum_{i=1}^{m} \left(y_i - \sum_{j=1}^{n} x_{ij}\beta_j\right)^2 + \lambda \sum_{j=1}^{n} |\beta_j| \tag{10}$$

- **HHO**

HHO is an optimization algorithm inspired by the hunting behavior of Harris hawks (Heidari, Mirjalili et al. 2019). In this algorithm, each hawk's fitness (i.e., fitness score) is evaluated using an objective function (e.g., classification accuracy in the study) of the optimization problem, denoted by $f(x)$. The algorithm incorporates social influences, where hawks share information on their positions and fitness scores to influence lower performers. A hawk's movement is determined by its own velocity ($v_i$) and position ($x_i$) relative to the current social hierarchy and its need to enhance its fitness score. The update equations for velocity and position are shown in Eq. 11 and Eq. 12, respectively

$$v_i(t+1) = wv_i(t) + c_1 r_1 \left(p_{i,\text{best}} - x_i(t)\right) + c_2 r_2 \left(g_{\text{best}} - x_i(t)\right) \tag{11}$$
$$x_i(t+1) = x_i(t) + v_i(t+1) \tag{12}$$

where $w$ is the inertial weight; $c_1$ and $c_2$ are the acceleration coefficients; $p_{i,\text{best}}$ is the best position of hawk $i$;

$g_{best}$ is the global best position of all hawks; $r_1$ and $r_2$ are random numbers between 0 and 1.

The social hierarchy of the hawks is determined by ranking them based on their fitness scores. Hawks with higher scores have higher social status. During the sharing phase, hawks exchange information on their positions and fitness scores to influence lower-ranked hawks. The sharing functions are given in Eq. 13 and Eq. 14.

$$z = f(x_i) + rand()\alpha\big(f(x_h) - f(x_l)\big) \tag{13}$$
$$x_i = x_i + rand()\beta(x_j - x_k) \tag{14}$$

where $z$ is the updated fitness score of a hawk; $\alpha$ and $\beta$ are weight factors; $rand()$ is a random number generator between 0 and 1; $f(x_h)$ and $f(x_l)$ are the fitness scores of the highest and lowest ranking hawks, respectively; $x_j$ is a hawk randomly selected from those sharing the same rank with the current hawk; $x_k$ is a hawk randomly selected from lower ranks.

The iterative process continues until a termination criterion is reached, such as the attainment of a predetermined fitness level or a maximum number of iterations. The final solution is the candidate solution with the highest fitness score.

- **Autoencoder model**

An autoencoder is usually a neural network designed to filter and to compress the representation of its input, and consists of two components: an encoder and a decoder. The encoder usually accepts a set of input data and compresses the information into intermediate vectors. The decoder is usually the prediction model. In our study, the decoder is an LSTM network, and the encoder is composed of LSTM and CNN layers.

Table. 2. Summary of the factor selection methods used

| Category | Explanation |
|---|---|
| Exhaustive method (benchmark) | exhaustive method (LR) |
| | exhaustive method (SVM) |
| | exhaustive method (RF) |
| | exhaustive method (Xgboost) |
| Filter method | IGR |
| Wrapper method | RFE-LR; RFE-SVM; RFE-RF; RFE-Xgboost; |
| Wrapper method | PSO-LR; PSO-SVM; PSO-RF; PSO-Xgboost; |
| Wrapper method | HHO-LR; HHO-SVM; HHO-RF; HHO-Xgboost; |
| Embedded method | LASSO-LR; LASSO-SVM; |
| Autoencoder method | LSTM-LSTM Encoder-Decoder model |
| Autoencoder method | CNN-LSTM Encoder-Decoder model |

*3.4. Generation of landslide susceptibility maps*

There are mainly two methods to produce the landslide susceptibility maps in the previous studies (Senouci, Taibi et al. 2021). In the first method, the susceptibility values (i.e., the probability of the landslide occurrence) at some random locations of the study area were first calculated, followed by a spatial interpolation (e.g., using inverse distance weight or IDW) to estimate the susceptibility values over the whole area (i.e., a susceptibility map). In the second method, the susceptibility values at all pixels of the study area were calculated directly. The choice of these two methods was mainly driven by the size of the study area and the computer performance. The second method was used in this study.

Based on the susceptibility values, each individual location/pixel of the susceptibility map was classified into one of the five classes using the Jenks natural break algorithm (Huang, Zhang et al. 2020): very high, high, moderate, low and very low.

## 4. Experiments details

The performances of all factor selection methods were tested using the Kerala dataset. The Kerala landslide map (i.e., Fig. 1) recorded a total number of 4728 identified landslides, each of which was represented by a pixel in the map. This map was used as the benchmark to check the LSM derived in this study. As the training of the models is based on a binary classification in which landslide and non-landslide samples were required, non-landslide samples in the same number as the identified landslides were randomly selected from the non-landslide area. Specifically, in order to ensure the spatial distribution of selected points is approximately uniform, we randomly generated 3-4 times more non-landslide points than landslide points in the study area, and the distance between each non-landslide point and its closest landslide point was filtered using a 1 km threshold. Furthermore, the total distance between each non-landslide point and its 5 nearest non-landslide points was calculated, and points with a total distance greater than 3 km were filtered out. Then, the Kerala dataset was randomly divided into three subsets for training (70%), validation (15%) and testing (15%), respectively. The training dataset was used to train the models considered, while the validation dataset was used to optimize the parameters of those models. Finally, the models were evaluated using the test dataset, the performances of which were compared. Five-fold cross-validation was used to objectively characterize the performances of the ML & DL models considered, and the mean value of the five-fold cross-validation results is used to represent model evaluation criteria.

### 4.1. Evaluation metrics

To assess the accuracy of the results produced by the models considered, the following set of evaluation criteria are considered in our investigation. True Positive (*TP*) and True Negative (*TN*) represent the number of correctly predicted landslide and non-landslide, respectively. False Positives (*FP*) and False Negatives (*FN*) represent the number of mispredicted landslides and non-landslide. Area Under Curve (*AUC*) representing the area under ROC (receiver operating characteristic) curve quantifies the prediction performance of a model, and its values range from 0.5 to 1. The higher the AUC value, the better the prediction performance of the model. In machine learning, the Kappa measures the degree of agreement between a pair of variables. Its values range from 0 to 1, where 1 indicates perfect agreement and perfect classification and 0 indicates inconsistency/independence, which are usually used as feed-forward neural networks. The statistical metrics by cross validation such as accuracy, *recall*, Precision, ROC curve and AUC, can be calculated using the following four metrics (i.e., *TP*, *TN*, *FP* and *FN*).

$$Accuracy = \frac{TP+TN}{TP+FP+TN+FN} \quad (10)$$

$$Recall = \frac{TP}{TP+FN} \quad (11)$$

$$Precision = \frac{TN}{TN+FP} \quad (12)$$

$$F1_{score} = 2 \times \frac{Precision \times Recall}{Precision+Recall} \quad (13)$$

$$Kappa = \frac{P_{Acc}-P_e}{1-P_e} \quad (14)$$

where, $P_{Acc}$ represents the classification accuracy; $P_e$ represents the random correct rate.

### 4.2. Configuration parameters

- **Factor selection parameters**

The factor selection criteria vary with the factor selection algorithms, which are controlled by their internal parameters (Shaheen, Agarwal et al. 2020). The parameters involved are introduced in Section 3.3. In PSO, $\omega$ keeps the particle moving with an inertia, giving it the tendency to expand the search space and the ability to explore new regions. $c_1$ and $c_2$ represent the weights of the statistical acceleration terms that push each particle towards the $\vec{p}(t)$ and $\vec{g}(t)$ positions. Lower values of $\omega$ allow the particles to hover outside the target region before being pulled back, and higher values cause the particles to rush abruptly towards or over the target region. Based on the empirical settings in the literature (Kalita and Singh 2020, Shaheen, Agarwal et al. 2020, Li, Xue et al. 2021, Zhou, Qiu et al. 2021), PSO is applied with the following configuration [$c_1$: 2, $c_2$: 2, $\omega$: 0.9], and its stopping condition is satisfied after 1000 iterations. For IGR, all factors attributable to the variable importance were ranked and the top 8 factors were selected based on the empirical importance threshold filtering (Tanyu, Abbaspour et al. 2021). RFE iteratively computes the cross-validation scores of all selections of contributing factors. The factors before the best cross-validation score are removed to obtain the best selection (Richhariya, Tanveer et al. 2020). The regularization parameter λ of the LASSO is 0.02, which is estimated by regularized path Least Absolute Shrinkage and Selection Operators using Cross-Validation (LASSOCV) by the Scikit-Learn module.

- **Network hyperparameters**

To minimize the influence of the model parameters on the final results, the hyperparameter of the ML models (LR, SVM, RF and Xgboost) in this study was optimized using the random-search-based parameter optimization. After the optimization, the hyperparameter of LR is set to be that the penalty factor is 100, and the loss function is optimized by the lib-linear solver. For SVM, the penalty factor *C* and gamma are set to 180 and 0.95, respectively, and the kernel is Radial Basis Function (RBF). For RF and Xgboost, *n_estimators* are set to 100 to avoid underfitting, and the maximum depth is 12. For the DL models, after establishing the basic framework, a random search method is used to optimize the number of units of the LSTM layer. All DL models' classifiers use two fully connected layers with 256 and 2 units, respectively. The final obtained model has 300 units of an LSTM encoder and a decoder LSTM with 100 units before the output layer. In the CNN-LSTM Encoder-Decoder model, the CNN encoder has two convolutional layers with one-dimensional convolutional kernels and kernel sizes [7, 1] and [3, 1], respectively. Different models' hyperparameters are listed in Table 3.

Table. 3. Summary of the model parameters used

| Model | Parameters range | Best Parameters |
|---|---|---|
| SVM | C = [0.1, 1, 10, 100, 180] | C = 180 |
| | degree = [1, ,2, 3, 4] | degree = 2 |
| | gamma = [0.01, 0.1, 0.5, 0.95] | gamma = 0.95 |
| | kernel = 'linear', 'poly', 'rbf' | kernel = 'rbf' |
| RF | n_estimators = [50,100,200] | n_estimators = 100 |
| | min_samples_split = [2,5,10,15] | min_samples_split = 10 |
| | min_samples_leaf = [2,5,10] | min_samples_leaf = 2 |
| | max_features = 'auto', 'sqrt', 'log2' | max_features = 'auto' |
| | max_depth = [2,5,12,15,20] | max_depth = 15 |
| | criterion = 'gini', 'entropy' | criterion = 'entropy' |
| LR | C = [0.1, 1, 10, 100, 180] | C = 100 |
| | penalty = 'l2' | penalty = 'l2' |
| | solver = 'lbfgs', 'liblinear' | solver = 'lib-linear' |
| Xgboost | n_estimators = [50,100,200] | n_estimators = 100 |
| | colsample_bytree = [0.1, 0.3, 0.5, 0.7, 0.9, 1.0] | colsample_bytree = 1.0 |
| | gamma = [0.01, 0.1, 0.5, 0.95] | gamma = 0.1 |
| | max_depth = [2,5,12,15,20] | max_depth = 12 |
| CNN | filters = [16, 32, 64, 128] | filters = 64 |
| | kernel_size = (3, 3), (5,5), (7,7) | kernel_size = (3, 3) |
| | dropout = [0.1, 0.2, 0.3] | dropout = 0.2 |
| | activation = 'relu', 'tanh' | activation = 'relu' |
| LSTM | units = [50, 100, 200, 300] | units = 100 |
| | dropout = [0.1, 0.2, 0.3] | dropout = 0.2 |
| | activation = 'relu', 'tanh' | activation = 'tanh' |
| LSTM-LSTM | LSTM1: units = [50, 100, 200, 300]; LSTM2:units = [50, 100, 200, 300] | LSTM1: units = 300; LSTM2: units = 100 |
| | dropout = [0.1, 0.2, 0.3] | dropout = 0.2 |
| | activation = 'relu', 'tanh' | activation = 'tanh' |
| CNN-LSTM | CNN: filters = [16, 32, 64, 128]; LSTM:units = [50, 100, 200, 300] | CNN: filters = 64; LSTM:units = 50 |
| | dropout = [0.1, 0.2, 0.3] | dropout = 0.3 |
| | kernel_size = (3, 3), (5,5), (7,7) | kernel_size = (3, 3) |
| | activation = 'relu', 'tanh' | activation = 'relu' |

## 5. Results

### 5.1. Exhaustive factor selection for the ML models

Using the exhaustive method, the performances of 32602 possible factor combinations, each containing a minimum of three factors, were evaluated under LR, SVM, RF and Xgboost classifiers were obtained, respectively. Fig. 5 shows the prediction accuracies of all the selections under each ML classifier. These accuracy values are grouped by the number of factors used in Fig. 5. The boxplot was adopted as it is a standardized way of displaying key statistics about the data distribution based on five-number summaries, including the minimum observed value (i.e., the lower edge or $Q_0$), the 25% quantile ($Q_1$), the median ($Q_2$), the 75% quantile ($Q_3$), and the maximum observed value (i.e., the upper edge or $Q_4$). The upper edge of the boxplot represents the optimal factor choice for maximum prediction accuracy.

It was observed in Fig. 5 that when more factors were used, the corresponding accuracy distribution was more concentrated towards the range of higher accuracy. It was also found that the maximum accuracy of individual groups tended to increase with the number of factors to a certain value before it started to decrease. This reflects that most of the factors considered were useful for the landslide prediction. Not only because LR was established by the linear relationship using all factors, but also because the loss function of LR was based on the maximum likelihood estimation function (Khammassi and Krichen 2017), poorly correlated or indistinguishable factors can lead to a more obvious decrease in the highest prediction accuracy of LR (shown in Fig. 5 (b)). This decrease started when the number of factors selected reached 12, indicating that approximately 4 factors had weaker correlations with landslide occurrences, and that all factors as the input were not the most effective selection.

The classification principle differs from one model to another model, making the influence of the factors on the prediction accuracy different. In Fig. 5, it was seen that the worst prediction accuracy (i.e., the lower edge) of individual group changed differently as the number of the input factors increased. Especially, the lower edge of LR did not change significantly as the number of the factors increased. As the LR model was based on a linear relationship and used all input factors for modeling, linearly indistinguishable factors could significantly affect the accuracy of the results (Engmann, Hart et al. 2009). Meanwhile, since LR assumed that the factors were independent of each other and ignored the interrelationship between the factors, the selection of insignificant factors or linearly indistinguishable factors would not increase the prediction performance. However, the SVM classifier did not rely on the overall data but used only the key support vector to determine the hyperplane for classification (Ding, Hua et al. 2014). The linearly indistinguishable factors were mapped to high-dimensional divisible spaces by kernel functions. This made SVM less influenced by the linear indistinguishable factors than LR. RF and Xgboost were nonlinear models developed on the decision tree, and were not affected by linearly indistinguishable factors.

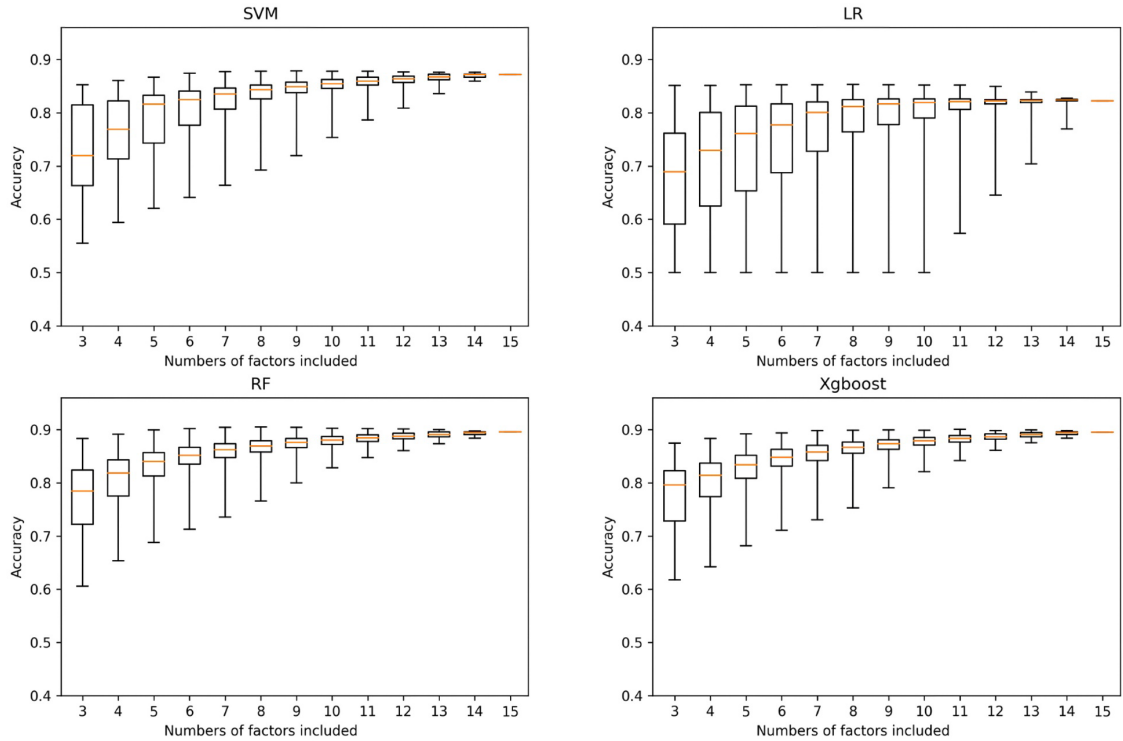

Fig. 5. The box plot of the prediction accuracies, grouped by the number of contributing factors used, for the exhaustive factor selection method: (a) SVM; (b) LR; (c) RF; (d) Xgboost.

Another useful analysis is to count the number of times each individual contributing factor was selected in each of the following four ranges: $Q_0 - Q_1$, $Q_1 - Q_2$, $Q_2 - Q_3$ and $Q_3 - Q_4$. This would indicate the relative importance of each factor as more important factors should be selected more times in the upper ranges (e.g., in $Q_3 - Q_4$). The numbers counted are shown in Fig. 6, where the number of times in those four ranges are shown in sequence by the bars from left to right, respectively, for each contributing factor. The taller the right-most bar for a factor is, the more contribution that factor would make to the prediction accuracy. As shown in Fig. 6, for different ML models, the key factors that were selected more times were not the same for the ML models considered. However, slope and elevation were selected as two key factors in all four models considered. In addition, distance-to-fault, distance-to-streams and rainfall were also identified as key factors in SVM, RF and Xgboost. Overall, there are clear similarities in the key factors as shown in Fig. 6 for SVM, RF and Xgboost, especially between RF and Xgboost. However, notable differences in key factors were found for LR.

More important factors would be selected more times from the range $Q_0 - Q_1$ to the range $Q_3 - Q_4$, such as rainfall, slope and elevation in Fig.7 (a, c, d). However, this was not always the case for LR (e.g., the range $Q_1 - Q_2$ for elevation in Fig.6 (b) is taller) for the following likely reason. Since LR is a generalized linear model, it is insufficient to utilize the relationships between factors. The selection of uncorrelated or linearly indistinguishable factors with significant factors (e.g., elevation) will affect the final prediction performance. As shown in Fig. 6 (b), elevation is influenced by insignificant factors, making it did not have the tendency of selected times to gradually increase.

Unlike LR, the kernel function mapping gave SVM a non-linear classification capability (Huang and Zhao 2018). RF and Xgboost were based on decision trees where the interrelationships of the multiple trees gave them a non-linear classification capability. In addition, RF and Xgboost were able to consider more interrelationships between factors as the tree depth increased. For these reasons, the key factors selected by SVM, RF and Xgboost, as indicated by increasing numbers of times each key factor was selected from the left-most bar to the right-most bar, were deemed to be more representative of the actual significance of those factors.

To further understand the significance of each factor, the prediction accuracies of all the 32767 selection cases were sorted in a descending order, based on which the top 1000 cases were used to count the times each factor was selected. A similar exercise was also done for the top 100 cases. These statistics are shown in Fig. 7, which suggest that distance-to-fault, distance-to-streams, rainfall, slope and elevation were more significant to the prediction accuracy.

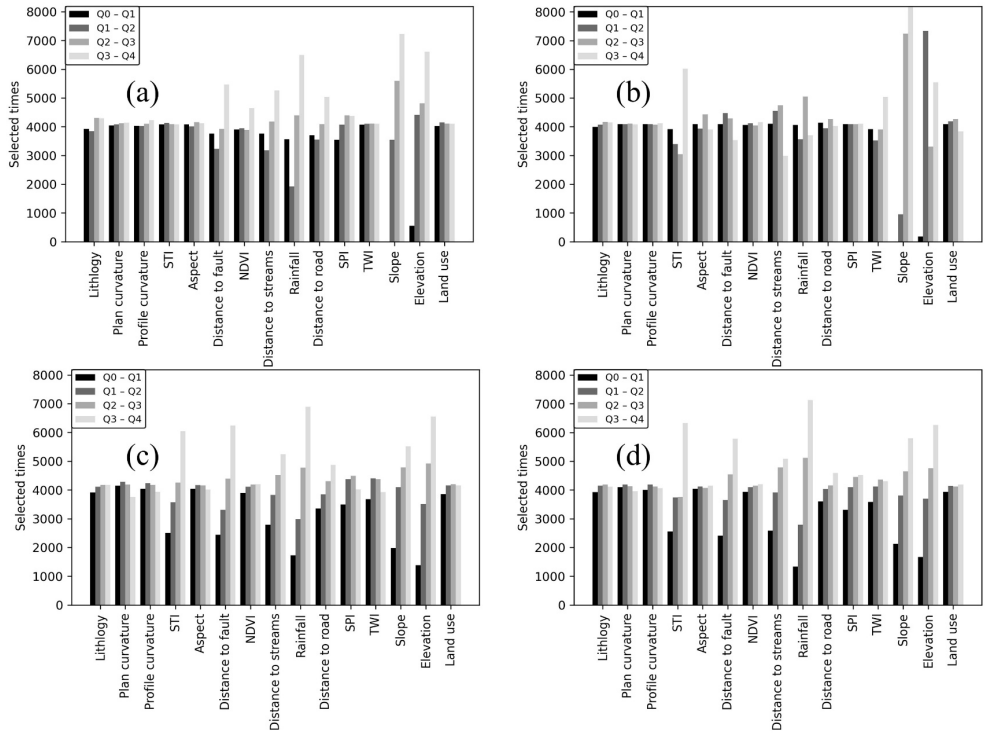

Fig. 6. Distributions of the number of times each factor was selected using the exhaustive method: (a) SVM; (b) LR; (c) RF; (d) Xgboost.

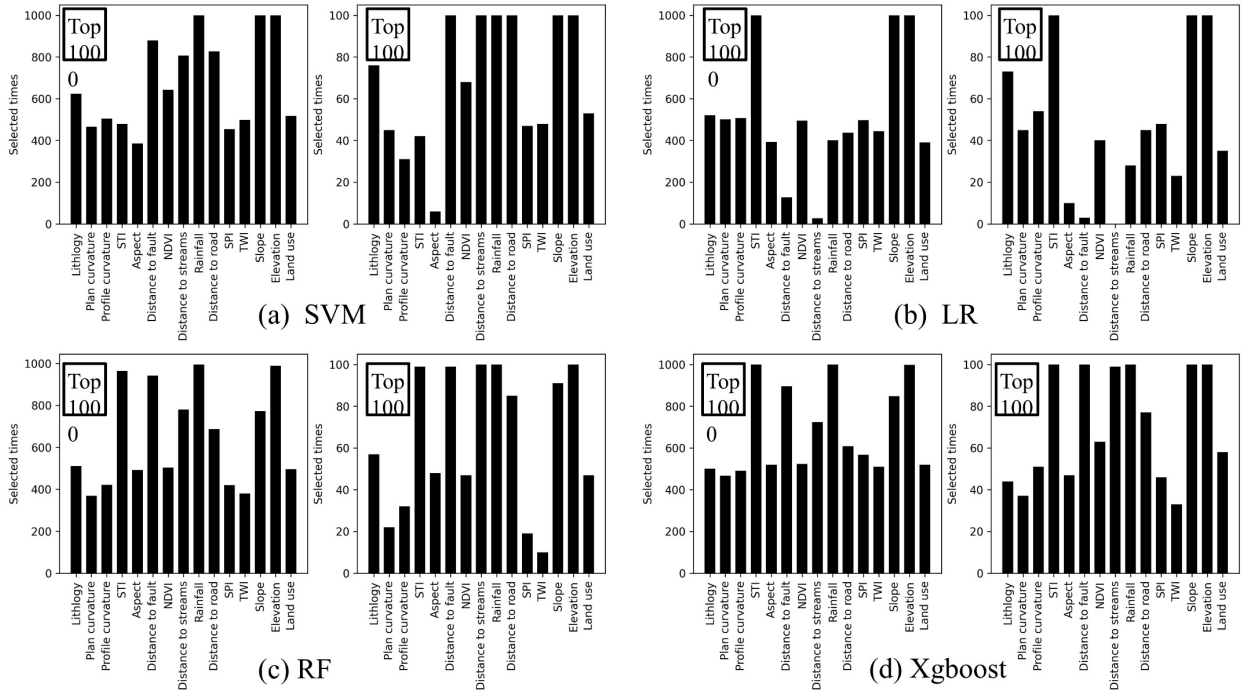

Fig. 7. Distribution statistics of the number of times each factor was selected in the selection cases of top accuracy using the exhaustive method: (a) SVM; (b) LR; (c) RF; (b) Xgboost.

Table 4 shows the factors selected in the optimal selection case of the highest prediction accuracy for each ML model, using the exhaustive method. In these optimal selections, rainfall, distance-to-road, slope and elevation were selected by all the models. Similar to the findings in Fig. 6 and Fig.7 where distance-to-fault, distance-to-streams, rainfall, distance-to-road, slope and elevation were more significant factors in SVM, RF and Xgboost, these factors were also selected as expected in the optimal selection cases. In addition, NDVI were also selected, although they were less significant in Fig. 6 and Fig. 7.

Table 4. The factors selected in the optimal selection from the exhaustive method for each ML model, where the factors selected are indicated by √, otherwise by ×.

| Factors | SVM | LR | RF | Xgboost |
|---|---|---|---|---|
| Lithology | √ | √ | × | √ |
| Plan | × | × | × | × |
| Profile | × | √ | × | √ |
| STI | × | √ | √ | √ |
| Aspect | × | × | × | × |
| Distance_to_fault | √ | × | √ | √ |
| NDVI | √ | × | √ | √ |
| Distance_to_stream | √ | × | √ | √ |
| **Rainfall** | √ | √ | √ | √ |
| **Distance_to_road** | √ | √ | √ | √ |
| SPI | × | √ | × | × |
| TWI | √ | × | × | × |
| **Slope** | √ | √ | √ | √ |
| **Elevation** | √ | √ | √ | √ |
| Landuse | × | × | √ | √ |

## 5.2. Comparisons of factor selection results for ML models

Table 5 shows the selection results obtained using the factor selection methods considered. It was observed that different selection mechanisms led to different choices, and so did different ML classifiers under the same selection mechanism. Based on the results in Table.5, rainfall, slope, and elevation were chosen as significant factors by all the selection methods, except distance to road which are consistent with the optimal selection shown in Table 4. In terms of the computational efficiency, LASSO was found to be much faster, while PSO and RFE were time-consuming because they required many iterations.

Table 5. The factor selection cases obtained using factor selection methods for the ML models

| Factor | IGR | RFE-SVM | RFE-LR | RFE-RF | RFE-Xgboost | PSO-SVM | PSO-LR | PSO-RF | PSO-Xgboost | HHO-SVM | HHO-LR | HHO-RF | HHO-Xgboost | LASSO-SVM | LASSO-LR |
|---|---|---|---|---|---|---|---|---|---|---|---|---|---|---|---|
| Lithology | × | √ | × | × | √ | × | × | × | √ | √ | × | √ | √ | √ | √ |
| Plan | × | √ | √ | × | × | × | × | × | × | × | × | × | √ | × | × |
| Profile | × | √ | √ | × | × | √ | × | × | × | √ | √ | × | √ | √ | × |
| STI | √ | √ | √ | √ | √ | × | × | × | × | × | √ | √ | × | × | √ |
| Aspect | × | × | √ | × | × | √ | √ | √ | √ | × | × | √ | √ | √ | √ |
| Distance to fault | √ | × | × | √ | √ | × | × | × | × | × | × | √ | × | √ | × |
| NDVI | × | √ | √ | × | × | √ | × | √ | √ | √ | √ | × | × | × | × |
| Distance to stream | √ | × | × | √ | √ | √ | √ | √ | √ | √ | × | × | × | × | × |
| **Rainfall** | √ | √ | √ | √ | √ | √ | √ | √ | √ | √ | × | × | √ | √ | √ |
| Distance to road | × | × | × | √ | × | √ | × | × | √ | √ | × | × | √ | √ | √ |
| SPI | × | √ | × | √ | × | × | × | × | × | × | √ | × | × | √ | √ |
| TWI | √ | √ | √ | √ | × | × | × | × | × | × | √ | √ | √ | × | √ |
| **Slope** | √ | √ | √ | √ | √ | √ | √ | √ | √ | × | √ | √ | × | √ | √ |
| **Elevation** | √ | √ | √ | √ | √ | √ | √ | √ | √ | √ | √ | √ | √ | √ | √ |
| Landuse | √ | × | × | × | × | × | × | × | × | × | √ | √ | √ | √ | √ |
| Total number of selected factors | 8 | 10 | 9 | 9 | 7 | 8 | 5 | 6 | 8 | 7 | 8 | 8 | 9 | 10 | 10 |
| Elapsed time (seconds) | 10.5 | 452.3 | 311.1 | 634.7 | 609.5 | 1253.2 | 1068 | 1636.1 | 1679.3 | 2437.1 | 752.8 | 924.6 | 2201.7 | 0.3 | 0.7 |

By comparing the results of the different factor selection method with different classifier, it can verify the advantages of selection method. Tables 6-9 show the performance matrices of LR, SVM, RF and Xgboost classifiers for various factor selection cases. Among them, LASSO-SVM, HHO-LR, HHO-RF, and PSO-Xgboost, marked with a bold font in Tables 6-9, achieved the best prediction accuracy with the utilization of their selected factors. It was found that these ML classifiers performed better when the factors selected were used in comparison to the case where all factors were used. The performances of RF and Xgboost were similar, and were better than those of SVM and LR. However, the improvements in the performances of these ML classifiers using the factors selected were not significant.

Table 6. The performance metrics of SVM under several factor selection cases

| Models | Selection cases | Testing accuracy | Precision | Recall | F1 score | AUC | Kappa |
|---|---|---|---|---|---|---|---|
| SVM | All factors | 0.8519 | 0.7970 | 0.9451 | 0.8648 | 0.8517 | 0.7037 |
| | The optimal case from the exhaustive method | 0.8752 | 0.8370 | 0.9324 | 0.8822 | 0.8751 | 0.7503 |
| | IGR | 0.8642 | 0.8246 | 0.9261 | 0.8724 | 0.8641 | 0.7284 |
| | RFE | 0.8681 | 0.8295 | 0.9275 | 0.8757 | 0.8680 | 0.7362 |
| | PSO | 0.8632 | 0.8222 | 0.9275 | 0.8717 | 0.8631 | 0.7263 |
| | HHO | 0.8575 | 0.8198 | 0.9160 | 0.8652 | 0.8576 | 0.7150 |
| | **LASSO** | **0.8688** | **0.8280** | **0.9317** | **0.8768** | **0.8687** | **0.7376** |

Table 7. The performance metrics of LR under several factor selection cases

| Models | | Testing accuracy | Precision | Recall | F1 score | AUC | Kappa |
|---|---|---|---|---|---|---|---|
| LR | All factors | 0.8240 | 0.8086 | 0.8501 | 0.8288 | 0.8240 | 0.6481 |
| | The optimal case from the exhaustive method | 0.8424 | 0.8238 | 0.8719 | 0.8472 | 0.8423 | 0.6847 |
| | IGR | 0.8318 | 0.8130 | 0.8628 | 0.8371 | 0.8317 | 0.6636 |
| | RFE | 0.8322 | 0.8186 | 0.8543 | 0.8361 | 0.8321 | 0.6643 |
| | PSO | 0.8343 | 0.8202 | 0.8571 | 0.8383 | 0.8342 | 0.6685 |
| | **HHO** | **0.8395** | **0.8280** | **0.8566** | **0.8421** | **0.8359** | **0.6790** |
| | LASSO | 0.8304 | 0.8100 | 0.8642 | 0.8362 | 0.8303 | 0.6607 |

Table 8. The performance metrics of RF under several factor selection cases

| Models | | Testing accuracy | Precision | Recall | F1 score | AUC | Kappa |
|---|---|---|---|---|---|---|---|
| RF | All factors | 0.8878 | 0.8804 | 0.9232 | 0.9013 | 0.8998 | 0.7977 |
| | The optimal case from the exhaustive method | 0.9073 | 0.8932 | 0.9253 | 0.9090 | 0.9078 | 0.8146 |
| | IGR | 0.9013 | 0.8856 | 0.9218 | 0.9033 | 0.9020 | 0.8026 |
| | RFE | 0.8999 | 0.8853 | 0.9190 | 0.9018 | 0.9005 | 0.7998 |
| | PSO | 0.8988 | 0.8845 | 0.9175 | 0.9007 | 0.8994 | 0.7977 |
| | **HHO** | **0.9032** | **0.8800** | **0.9353** | **0.9068** | **0.9033** | **0.8065** |

Table 9. The performance metrics of Xgboost under several factor selection cases

| Models | | Testing accuracy | Precision | Recall | F1 score | AUC | Kappa |
|---|---|---|---|---|---|---|---|
| Xgboost | All factors | 0.8938 | 0.8851 | 0.9281 | 0.9061 | 0.9047 | 0.8075 |
| | The optimal case from the exhaustive method | 0.9094 | 0.8936 | 0.9295 | 0.9112 | 0.9101 | 0.8188 |
| | IGR | 0.9059 | 0.8866 | 0.9309 | 0.9082 | 0.9069 | 0.8118 |
| | RFE | 0.9031 | 0.8891 | 0.9211 | 0.9048 | 0.9036 | 0.8061 |
| | **PSO** | **0.9076** | **0.8906** | **0.9295** | **0.9097** | **0.9084** | **0.8153** |
| | HHO | 0.9035 | 0.8858 | 0.9245 | 0.9047 | 0.9046 | 0.8070 |

Table 10 shows the performance metrics of the ML classifiers using the significant factors that were identified by all the classifiers in the exhaustive method, which include elevation, slope, distance-to-road and rainfall. The results show that using these three factors led to the performances that were close to those where all factors were used. As shown in Fig. 5, using a small number of factors does not guarantee the stability of the model prediction.

Table 10. The performance metrics of the models using the most significant factors

| Models | Testing accuracy | Precision | Recall | F1 score | AUC | Kappa |
|---|---|---|---|---|---|---|
| SVM | 0.8589 | 0.8081 | 0.9310 | 0.8686 | 0.8588 | 0.7178 |
| LR | 0.8311 | 0.8266 | 0.8388 | 0.8326 | 0.8310 | 0.6621 |
| RF | 0.8794 | 0.8609 | 0.9057 | 0.8827 | 0.8794 | 0.7588 |
| Xgboost | 0.8805 | 0.8597 | 0.9099 | 0.8841 | 0.8804 | 0.7609 |

The results in the Table 6-9 show that for the ML models, most of the factor selection methods achieved better results than the case of all factors being the input. The performances were even close to the best performance in some cases, but varied greatly between the models. The improvement in prediction accuracy from a suitable model is much greater than the factor selection. Therefore, for LSM, it is important to select an appropriate model to get more stable and accurate results. In this study, due to the nonlinear modelling capabilities of RF and Xgboost, they showed better prediction accuracy.

*5.3. Factor selection results for DL-based model*

Table 11 shows the performance results of the DL models considered. Some previous studies used the filter method, or the wrapper method to select factors for DL models (Wang, Fang et al. 2019, Sameen, Pradhan et al. 2020, Hakim, Rezaie et al. 2022). We compared some representative factor selection methods in the literature for the DL models, including the IGR (Wang, Fang et al. 2019), RFE-SVM (Hong, Adler et al. 2007) and autoencoder (Li and Becker 2021). The case that all factors were used was the baseline.

In terms of the performances of DL models under the baseline case, LSTM performed the worst and the CNN-LSTM Encoder-Decoder model achieved the best performance with a final test accuracy of 93.23%. The performance of CNN was between LSTM and the CNN-LSTM Encoder-Decoder model. In terms of the matrices recall and precision, the values of recall in Tables 6-9 were greater than those of precision, which suggests that the prediction accuracy for landslide events is lower than that for non-landslide events. In Table 10, it is found that the values of recall and precision were closer in most of the models. These suggested that the DL models predicted landslide events more accurately than the ML models considered.

For all the DL models tested, the baseline case outperformed the cases of using selected factors. This indicates that the factor selection was not an effective means of improving the performance of the DL models. CNN and LSTM had mechanisms embedded in their network structures for extracting contributing factors (features), unlike the extraction of specific factors by ML [45]. The DL models integrated the feature extracting and the learning process, so that the use of the traditional factor selection approaches was not desirable for the DL models. For the DL models, the encoder was used to select contributing factors by reinforcing the channel weights in the networks. Due to the end-to-end characteristics of the DL models, it would be better to rely on the network's own weights to determine the importance of individual contributing factors compared to the traditional factor selection, as suggested by the performance metrics shown in Table 11.

The autoencoder with an encoding-decoding mechanism enables the model to learn to update the weights so that the factor extraction favored more important factors. It is evident from the performance metrics in Table 11 that the autoencoder structure improved the prediction accuracy of the DL models.

Table 11. The performance metrics of the DL models

| Models | Factor selection | Training accuracy | Testing accuracy | AUC | Precision | Recall | F1 score | Kappa |
|---|---|---|---|---|---|---|---|---|
| CNN (DL) | ALL | 0.9126 | 0.9078 | 0.9218 | 0.9021 | 0.9101 | 0.9061 | 0.8211 |
| | IGR | 0.8615 | 0.8435 | 0.8684 | 0.8182 | 0.8482 | 0.8329 | 0.7813 |
| | RFE-SVM | 0.8790 | 0.8554 | 0.8876 | 0.8526 | 0.8564 | 0.8545 | 0.7434 |
| LSTM (DL) | ALL | 0.8739 | 0.8692 | 0.9076 | 0.8670 | 0.8710 | 0.8690 | 0.8115 |
| | IGR | 0.8530 | 0.8413 | 0.8629 | 0.8413 | 0.8413 | 0.8413 | 0.7732 |
| | RFE-SVM | 0.8563 | 0.8458 | 0.8640 | 0.8466 | 0.8466 | 0.8466 | 0.7132 |
| LSTM-LSTM Encoder-Decoder model (DL) | ALL | 0.9233 | 0.9210 | 0.9320 | 0.9178 | 0.9243 | 0.9210 | 0.8256 |
| CNN-LSTM Encoder-Decoder model (DL) | ALL | 0.9477 | 0.9323 | 0.9356 | 0.9317 | 0.9346 | 0.9331 | 0.8406 |

*5.4. Model performance comparison using Wilcoxon Tests*

Although DL models such as CNN-LSTM Encoder-Decoder have demonstrated sufficient predictive accuracy in LSM, further investigation is needed to analyze the impact of factor selection and model selection on establishing an appropriate LSM model. Therefore, a non-parametric pairwise signed-rank test (such as the Wilcoxon test) is suitable for evaluating the statistical significance of performance differences between different LSM models using different factor selection methods (Merghadi, Yunus et al. 2020). For pairwise comparisons between different models or different factor selection methods using the Wilcoxon test, a p-value less than or equal to 0.05 indicates that there is a significant difference in model performance, while a p-value greater than 0.05 indicates that the models are similar in performance. The comparison of model performance involves conducting Wilcoxon tests on the accuracy metrics obtained during both the hyperparameter optimization process and the factor selection optimization process, as these processes affect model performance.

- **Hyperparameter optimization process**

Fig. 8 displays the boxplot of the accuracy distribution during hyperparameter optimization of various LSM models considered. These models exhibited varying ranges of accuracy during the hyperparameter optimization. SVM, RF, and CNN showed relatively larger variations, which suggests that SVM, RF and CNN were more susceptible to the impact of hyperparameters within the hyperparameter range shown in Table 3. In contrast, the variation ranges of LR, XGBoost, LSTM, LSTM-LSTM Encoder-Decoder and CNN-LSTM Encoder-Decoder were found to be relatively small. In combination with the accuracy values reported in Tables 6-10, some LSM models (e.g., Xgboost, LSTM, LSTM-LSTM Encoder-Decoder and CNN-LSTM Encoder-Decoder) were minimally influenced by hyperparameter optimization and can deliver superior performance. The computing time of the hyperparameter optimization of each LSM model is shown in Table 12.

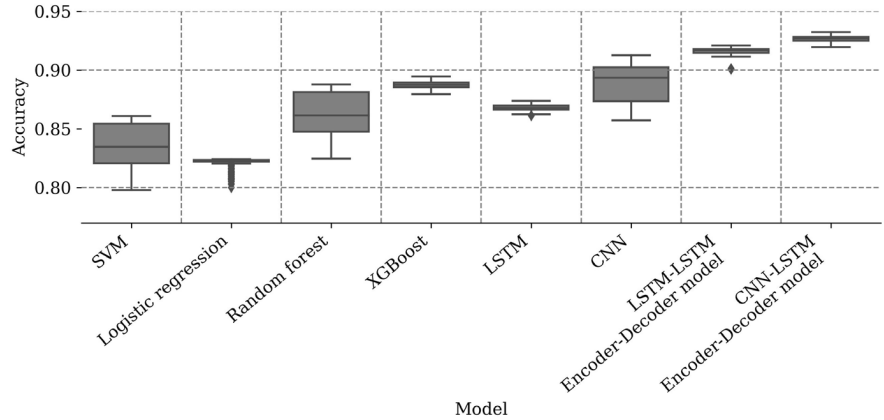

Fig. 8. Boxplot of the accuracy distribution during hyperparameter optimization of different LSM models.

Table 12. Computing time of the hyperparameter optimization of different LSM models.

| Models | SVM | LR | RF | Xgboost | LSTM | CNN | LSTM-LSTM Encoder-Decoder | CNN-LSTM Encoder-Decoder |
|---|---|---|---|---|---|---|---|---|
| Elapsed time (seconds) | 6211.3 | 1453.0 | 5757.1 | 1179.7 | 3897.2 | 3360.5 | 4908.4 | 4533.3 |

- **Factor selection optimization process**

Fig. 9 shows the boxplots illustrating the distribution of accuracy for various LSM models during factor selection. The accuracy values depicted in Fig. 9 were derived from applying factor combinations (shown in Table.5) to six randomly selected subsets of data. The testing accuracy matrix effectively facilitates the evaluation and comparison of the generalization of different factor selection methods. Notably, Fig. 10 shows that HHO attained the smallest range of accuracy among all methods. Furthermore, all factor selection methods demonstrated enhanced generalization performance when all factors were provided as input.

The statistical results shown in Fig. 10 clearly suggest that the factor selection methods had significant impacts on the performances of SVM, LR and RF models, compared to using all factors as input. However, for the Xgboost model, only HHO and IGR methods exhibited significant improvements through factor selection. Nevertheless, these findings were deemed adequate to demonstrate that using factor selection methods can significantly improve model performance, compared to not using such factor selection methods.

Furthermore, the best performing models for ML and DL were statistically tested for significance by comparing their performance against six randomly selected subsets of the data, as shown in Figure 11. It was observed that all pairwise comparisons between the models had a p-value less than 0.05, indicating significant differences in their performance. Overall, based on the analysis of prediction accuracies and significance tests, the CNN-LSTM Encoder-Decoder model exhibited the best performance among all proposed LSM models and showed clear advantages over other models.

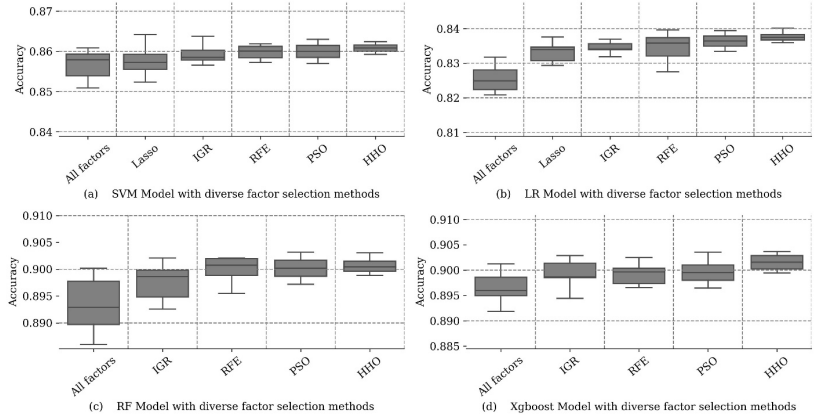

Fig. 9. Boxplots of the accuracy distributions for different factor selection methods and different LSM models.

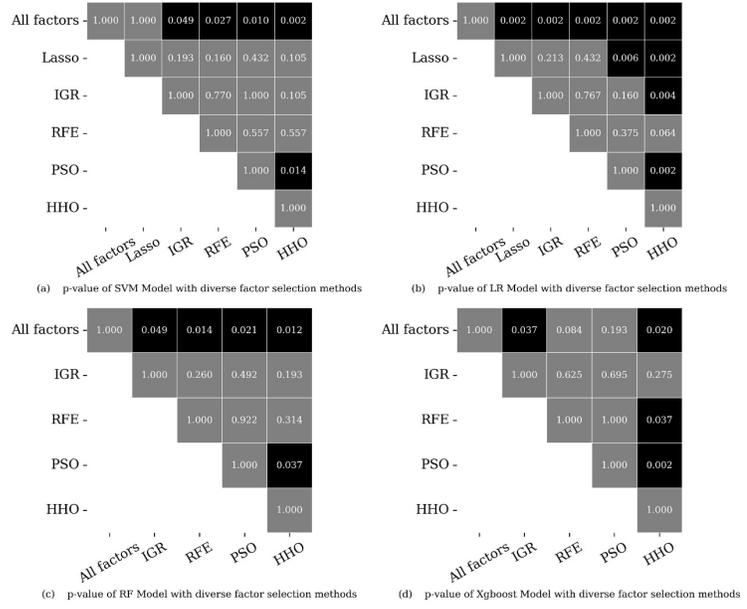

Fig. 10. Wilcoxon tests results for comparison between different factor selection methods (significance achieved at p <0.05 in black color)

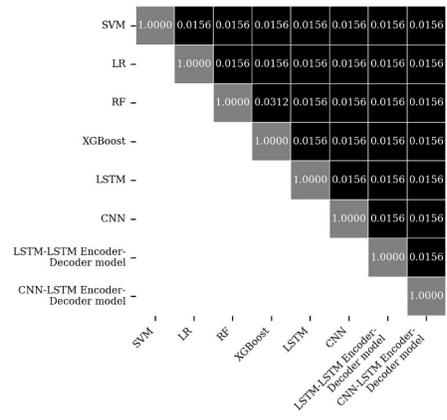

Fig. 11. Wilcoxon Tests results for LSM methods comparison after hyperparameter optimization with all factors as inputs (significance achieved at $p < 0.05$ in black color)

## 6. Landslide susceptibility mapping

The LSM outputs from different classifier models (i.e., SVM, LR, RF, Xgboost and CNN-LSTM Encoder-Decoder model) using the associated optimal selection case are shown in Fig.12. Each pixel of the susceptibility maps belongs to one of the five risk categories: very high, high, moderate, low and very low. Two local zones (as shown in Fig. 1) were also zoomed in to clearly visualize the risk categories against the individual historical landslides recorded. Recorded historical landslides can be used to verify the quality of a landslide susceptibility map (Wang, Zhang et al. 2021). In the landslide susceptibility maps shown in Fig. 12, the historical landslide locations fell well within high or very high risk areas. In addition, the boundaries between landslide areas and non-landslide areas in those maps were clear.

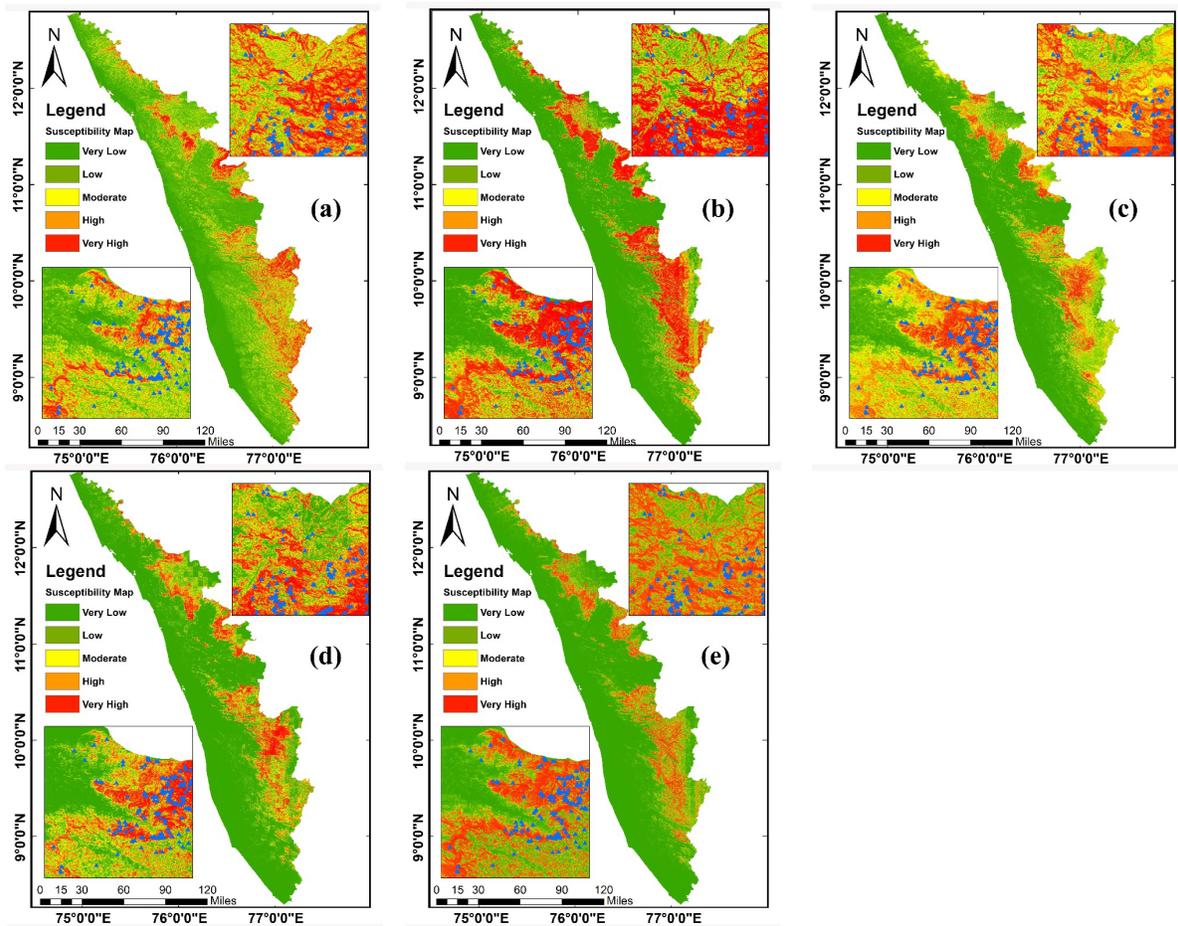

Fig. 12. Comparison of LSM outputs derived from different ML and DL models using the optimal selection case: (a) LR, (b) SVM, (c) RF, (d) Xgboost, (e) CNN-LSTM Encoder-Decoder.

To quantitatively compare the prediction accuracies of LSM derived using the different classifier models, the ratios of non-landslide pixels to all pixels of each susceptibility class (e.g., Fig. 13 (a)), and the ratios of historical landslide pixels to all pixels of each susceptibility class (e.g., Fig. 13 (b)) using the optimal selection case were calculated and compared. For more accurate LSM, the followings are expected: a higher percentage in the "very low" class and a lower percentage in the "very high" class in Fig. 13 (a); a lower percentage in the "very low" class and a higher percentage in the "very high" class in Fig.13 (b).

In Fig.13 (a), as anticipated, the percentages of LR, SVM, RF, Xgboost and CNN-LSTM Encoder-Decoder in the "very low" class increased sequentially while the percentages in the "very high" class declined sequentially. In Fig. 13 (b), the percentages in the "very high" class grew from LR to CNN-LSTM Encoder-Decoder, whilst the percentages in the "very low" class declined. For the study data considered, the DL model (CNN-LSTM Encoder-Decoder) outperformed the ML models used, and meanwhile the ML model Xgboost achieved the highest prediction accuracy amongst the ML models used. These demonstrated that choosing a more appropriate model for LSM can improve the LSM accuracy more significantly than using factor selection.

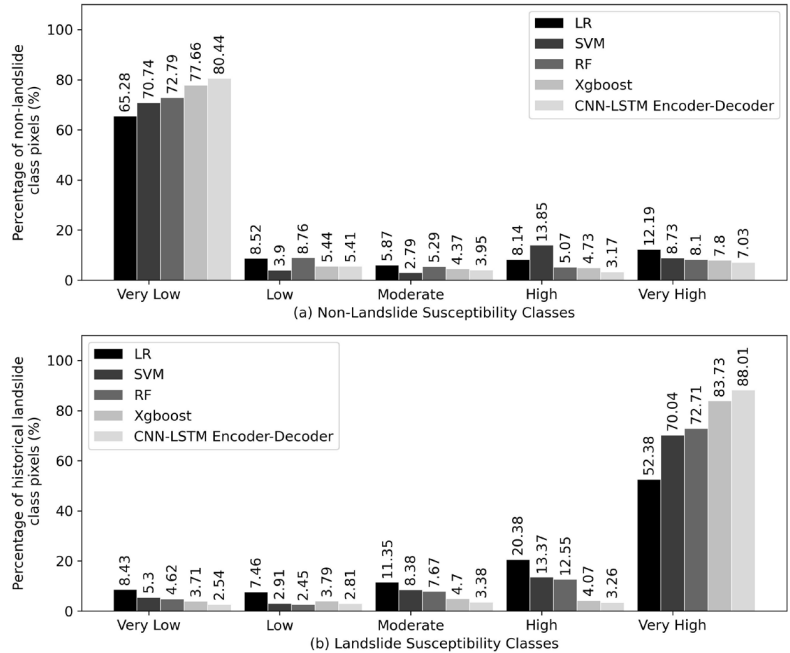

Fig. 13. Statistics of non-landslides and landslides (as recorded in the landslide inventory) located in each susceptibility class of the landslide susceptibility maps for different classifier models: (a) the percentage of non-landslide pixels, (b) the percentage of landslide pixels.

## 7. Discussion

The discussion is focused on the key issues regarding the use of ML and DL models in LSM. Specifically, the follow aspects are discussed, including uncertainties in the data used in employed models, limitations of the hyperparameter optimization process, and comparisons between different ML and DL model frameworks. Through our investigations, a set of valuable recommendations are provided guide the design and adoption of ML and DL models in LSM. Additionally, this discussion provides an overview of the limitations of this study, which facilitate the development of future research initiatives in this field.

### 7.1. Uncertainties of the data used in the model.

Because the ML and DL models considered in this study were data-driven, the quality of the contributing factor maps can affect the model performance. The measurement accuracy of the data would affect the quality of these maps, which depends mainly on the associated instruments and the platforms/users supporting the instruments. In our study, a range of contributing factors were considered. It was unlikely that all the information associated with these contributing factors were collected at the same time. It was more likely that there were temporal differences in the information collected. These might lead to inconsistency of the information used for the modeling if temporal changes in those contributing factors occurred, which would subsequently affect the prediction accuracy of landslide susceptibility maps. In addition, the data used to derive the contributing factors often have varying spatial resolutions. To produce the contributing factor maps of the same spatial resolution, interpolation is required, which may lead to inconsistent local spatial variations between the contributing factor maps.

The data of some contributing factors were acquired by the instruments from different platforms. These data often needed to be transformed into the same coordinate system. The transformation of coordination systems would inevitably generate positional errors. These errors may not only lead to local inconsistency of information between the contributing factors, but also cause mismatches between the landslide positions in the landslide inventory and the corresponding pixels in the contributing factor maps.

The construction of landslide susceptibility maps using ML and DL models relied heavily on the credibility of a historical landslide inventory. However, the inventory may not record all the landslide cases. In addition, as the historical landslides were often recorded manually, they were represented by single point locations in the map instead of their full spatial extents (e.g., the landslide inventory used in our study). The spatial extents of landslides (especially large ones) would be preferred to train learning models more accurately. This requires adequate field surveys or fine-resolution remote sensing to obtain a more detailed characterization of landslides.

*7.2. Limitations of the hyperparameter optimization process*

The hyperparameters of ML and DL models can affect the prediction performance of the models. For a given classifier, its optimal hyperparameters vary with the input conditioning factors. To determine the optimal selection of factors, the determination of hyperparameters was conducted by random search using all factors considered to make them favorable towards the potential optimal selection(s) in this study. However, it should be noted that the hyperparameters cannot exhaustively be optimized for every possible selection of factors due to the limitation of computing capability in this study.

*7.3. Comparison between different model frameworks*

This study provides a comprehensive overview of the application of mainstream ML and DL models to LSM. We also optimized the models' hyperparameters and performed factor selection to enhance model performance. Overall, based on the analysis of predictive accuracy and significance testing, the CNN-LSTM Encoder-Decoder model exhibited superior performance compared to other models considered. In contrast, the linear model LR consistently produced the lowest accuracy results among all implemented models. The Xgboost model provided similar or slightly better performance than the RF model. However, it should be noted that there was no single or specific model that can be applied to all LSM scenarios (Merghadi, Yunus et al. 2020). The suitability of a model for real-world problems depends on a variety of considerations, including computational efficiency, prediction accuracy, computational constraints and user interfaces.

LR is easy to use, requires minimal computational resources, and has smaller hyperparameters. However, it cannot solve complex nonlinear relationships because the decision boundary is linear. When there are multiple or nonlinear decision boundaries, the LR model often performs poorly. Additionally, factor selection plays a crucial role in LR model performance, as certain nonlinear or collinear features can affect the model performance. SVM specifies different kernel functions for the decision function and thus has good nonlinear processing ability. However, it is limited by the kernel function, which determines the computational complexity and nonlinear processing ability. Furthermore, SVM is sensitive to outliers and prone to overfitting when trained on small samples with outlier data. RF and Xgboost are robust to noisy data such as outliers and can also provide highly accurate performance, but can overfit when dealing with small datasets or data with strong linear relationships. CNN, LSTM, and their derivative models are more suitable for large-scale data, and due to their more trainable parameters, they exhibit superior performance in complex scenarios, enabling them to discover more potential relationships between factors and landslide events. However, complex network architectures result in increased computational burden.

In summary, DL models outperform SVM, RF, and LR models in terms of accuracy. However, they require more computational resources, which may pose challenges for users with limited resources. Furthermore, working with DL models requires some skill, as incorrect hyperparameter settings can affect model performance. Automated hyperparameter tuning techniques, such as random search, eliminate the need for extensive human involvement in the modeling process. Moreover, since GPUs provide several times faster than CPUs for DL computations (Biswas, Kuppili et al. 2019), DL models are not slower than SVM and RF models in practical use of LSM. Ultimately, the choice of an appropriate LSM model depends on the users' expectations for LSM output, their understanding of ML and DL models, and their available resources. For a balance between fast results and ease of implementation, SVM, LR, or Xgboost can be considered, which typically provide good predictive performance and have limited parameters that are easy to understood and optimize. More complex DL models can provide more accurate performance if users have sufficient experience with these models.

## 8. Conclusion

In this study, four typical factor selection methods (i.e., IGR, RFE, PSO and LASSO) for landslide susceptibility mapping were investigated and compared for four typical ML models (i.e., SVM, LR, RF and Xgboost). An exhaustive method was used to search for the optimal case of factor selection as a baseline. It was found that the prediction accuracy of the ML models tended to increase with the number of factors selected to a certain value before it started to decrease. This decrease was likely caused by the inclusion of less important factors. The most significant factors identified by all the selection methods include rainfall, slope and elevation. All the factor selection methods improved the prediction accuracy of the ML models in comparison to the case where all contributing factors were used. However, none of these selection methods could select the optimal set of factors determined by the exhaustive method. The contributing factors selected under different combinations of ML models and selection methods (except IGR) varied, suggesting that the factor selection was affected not only by the factor selection methods but also by the ML algorithms. Based on the prediction performances obtained, the choice of ML models can provide greater improvement than the factor selection. Amongst the ML models considered, Xgboost outperformed the other models.

For the DL models (i.e., CNN and LSTM), two factor selection methods (i.e., RFE-SVM and IGR) were considered. Contrary to the case of ML, these conventional factor selection methods reduced the prediction accuracy of the DL models. Incorporating an autoencoder with an encoding-decoding mechanism into the DL models led to higher prediction accuracies than SVM, LR, RF and Xgboost.

**Statements and Declarations**

*Funding*

This research was funded by Xi'an Jiaotong-Liverpool University Key Program Special Fund under grant number KSF-E-40 and Research Enhancement Fund under grant number REF-21-01-003.

*Conflict of Interests*

The authors declare no competing interests.

*Compliance with Ethical Standards*

**Reference**

Acharya, T. D. (2018). Regional scale landslide hazard assessment using machine learning methods in Nepal, PhD Thesis, Kangwon National University, Chuncheon, Korea.

Aleotti, P. and R. Chowdhury (1999). "Landslide hazard assessment: summary review and new perspectives." Bulletin of Engineering Geology and the environment **58**(1): 21-44.

Biswas, M., V. Kuppili, L. Saba, D. R. Edla, H. S. Suri, E. Cuadrado-Godia, J. R. Laird, R. T. Marinhoe, J. M. Sanches and A. Nicolaides (2019). "State-of-the-art review on deep learning in medical imaging." Frontiers in Bioscience-Landmark **24**(3): 380-406.

Bordoni, M., C. Meisina, R. Valentino, N. Lu, M. Bittelli and S. Chersich (2015). "Hydrological factors affecting rainfall-induced shallow landslides: from the field monitoring to a simplified slope stability analysis." Engineering Geology **193**: 19-37.

Can, R., S. Kocaman and C. Gokceoglu (2021). "A comprehensive assessment of XGBoost algorithm for landslide susceptibility mapping in the upper basin of Ataturk dam, Turkey." Applied Sciences **11**(11): 4993.

Ding, S., X. Hua and J. Yu (2014). "An overview on nonparallel hyperplane support vector machine algorithms." Neural computing and applications **25**(5): 975-982.

Dou, J., A. P. Yunus, D. T. Bui, A. Merghadi, M. Sahana, Z. Zhu, C.-W. Chen, Z. Han and B. T. Pham (2020). "Improved landslide assessment using support vector machine with bagging, boosting, and stacking ensemble machine learning framework in a mountainous watershed, Japan." Landslides **17**: 641-658.
Engmann, S., B. M. Hart, T. Sieren, S. Onat, P. König and W. Einhäuser (2009). "Saliency on a natural scene background: Effects of color and luminance contrast add linearly." Attention, Perception, & Psychophysics **71**(6): 1337-1352.
Fang, Z., Y. Wang, L. Peng and H. Hong (2021). "Predicting flood susceptibility using LSTM neural networks." Journal of Hydrology **594**: 125734.
Fiorucci, F., F. Ardizzone, A. C. Mondini, A. Viero and F. Guzzetti (2019). "Visual interpretation of stereoscopic NDVI satellite images to map rainfall-induced landslides." Landslides **16**(1): 165-174.
Gaidzik, K. and M. T. Ramírez-Herrera (2021). "The importance of input data on landslide susceptibility mapping." Scientific Reports **11**(1): 19334.
Ghosh, P., S. Azam, M. Jonkman, A. Karim, F. J. M. Shamrat, E. Ignatious, S. Shultana, A. R. Beeravolu and F. De Boer (2021). "Efficient prediction of cardiovascular disease using machine learning algorithms with relief and LASSO feature selection techniques." IEEE Access **9**: 19304-19326.
Gorelick, N., M. Hancher, M. Dixon, S. Ilyushchenko, D. Thau and R. Moore (2017). "Google Earth Engine: Planetary-scale geospatial analysis for everyone." Remote sensing of Environment **202**: 18-27.
Hakim, W. L., F. Rezaie, A. S. Nur, M. Panahi, K. Khosravi, C.-W. Lee and S. Lee (2022). "Convolutional neural network (CNN) with metaheuristic optimization algorithms for landslide susceptibility mapping in Icheon, South Korea." Journal of Environmental Management **305**: 114367.
Hao, L., A. Rajaneesh, C. Van Westen, K. Sajinkumar, T. R. Martha, P. Jaiswal and B. G. McAdoo (2020). "Constructing a complete landslide inventory dataset for the 2018 monsoon disaster in Kerala, India, for land use change analysis." Earth system science data **12**(4): 2899-2918.
Heidari, A. A., S. Mirjalili, H. Faris, I. Aljarah, M. Mafarja and H. Chen (2019). "Harris hawks optimization: Algorithm and applications." Future generation computer systems **97**: 849-872.
Hong, Y., R. F. Adler, A. Negri and G. J. Huffman (2007). "Flood and landslide applications of near real-time satellite rainfall products." Natural Hazards **43**(2): 285-294.
Huang, F., J. Zhang, C. Zhou, Y. Wang, J. Huang and L. Zhu (2020). "A deep learning algorithm using a fully connected sparse autoencoder neural network for landslide susceptibility prediction." Landslides **17**(1): 217-229.
Huang, Y. and L. Zhao (2018). "Review on landslide susceptibility mapping using support vector machines." Catena **165**: 520-529.
Kalita, D. J. and S. Singh (2020). "SVM hyper-parameters optimization using quantized multi-PSO in dynamic environment." Soft Computing **24**(2): 1225-1241.
Khammassi, C. and S. Krichen (2017). "A GA-LR wrapper approach for feature selection in network intrusion detection." computers & security **70**: 255-277.
Kim, J.-C., S. Lee, H.-S. Jung and S. Lee (2018). "Landslide susceptibility mapping using random forest and boosted tree models in Pyeong-Chang, Korea." Geocarto international **33**(9): 1000-1015.
Kotsiantis, S. (2011). "Feature selection for machine learning classification problems: a recent overview." Artificial Intelligence Review **42**(1): 157-176.
Kuriakose, S. L., G. Sankar and C. Muraleedharan (2009). "History of landslide susceptibility and a chorology of landslide-prone areas in the Western Ghats of Kerala, India." Environmental geology **57**(7): 1553-1568.
Lee, S. and J. A. Talib (2005). "Probabilistic landslide susceptibility and factor effect analysis." Environmental Geology **47**(7): 982-990.
Leynaud, D., T. Mulder, V. Hanquiez, E. Gonthier and A. Régert (2017). "Sediment failure types, preconditions and triggering factors in the Gulf of Cadiz." Landslides **14**(1): 233-248.
Li, A.-D., B. Xue and M. Zhang (2021). "Improved binary particle swarm optimization for feature selection with new initialization and search space reduction strategies." Applied Soft Computing **106**: 107302.
Li, W. and D. M. Becker (2021). "Day-ahead electricity price prediction applying hybrid models of LSTM-based deep learning methods and feature selection algorithms under consideration of market coupling." Energy **237**: 121543.
Lin, G.-F., M.-J. Chang, Y.-C. Huang and J.-Y. Ho (2017). "Assessment of susceptibility to rainfall-induced landslides using improved self-organizing linear output map, support vector machine, and logistic regression." Engineering geology **224**: 62-74.
Long, W., J. Jiao, M. Xu, M. Tang, T. Wu and S. Cai (2022). "Lens-imaging learning Harris hawks optimizer for global optimization and its application to feature selection." Expert Systems with Applications **202**: 117255.
Malekipirbazari, M., V. Aksakalli, W. Shafqat and A. Eberhard (2021). "Performance comparison of feature selection and extraction methods with random instance selection." Expert Systems with Applications **179**: 115072.
Marjanović, M., M. Kovačević, B. Bajat and V. Voženílek (2011). "Landslide susceptibility assessment using SVM

machine learning algorithm." Engineering Geology **123**(3): 225-234.

Merghadi, A., A. P. Yunus, J. Dou, J. Whiteley, B. ThaiPham, D. T. Bui, R. Avtar and B. Abderrahmane (2020). "Machine learning methods for landslide susceptibility studies: A comparative overview of algorithm performance." Earth-Science Reviews **207**: 103225.

Micheletti, N., L. Foresti, S. Robert, M. Leuenberger, A. Pedrazzini, M. Jaboyedoff and M. Kanevski (2014). "Machine learning feature selection methods for landslide susceptibility mapping." Mathematical geosciences **46**(1): 33-57.

Mugagga, F., V. Kakembo and M. Buyinza (2012). "Land use changes on the slopes of Mount Elgon and the implications for the occurrence of landslides." Catena **90**: 39-46.

Ohlmacher, G. C. (2007). "Plan curvature and landslide probability in regions dominated by earth flows and earth slides." Engineering Geology **91**(2-4): 117-134.

Peduto, D., M. Santoro, L. Aceto, L. Borrelli and G. Gullà (2021). "Full integration of geomorphological, geotechnical, A-DInSAR and damage data for detailed geometric-kinematic features of a slow-moving landslide in urban area." Landslides **18**(3): 807-825.

Pham, B. T., D. Van Dao, T. D. Acharya, T. Van Phong, R. Costache, H. Van Le, H. B. T. Nguyen and I. Prakash (2021). "Performance assessment of artificial neural network using chi-square and backward elimination feature selection methods for landslide susceptibility analysis." Environmental Earth Sciences **80**(20): 1-13.

Ramasamy, S., S. Gunasekaran, J. Saravanavel, R. M. Joshua, R. Rajaperumal, R. Kathiravan, K. Palanivel and M. Muthukumar (2021). "Geomorphology and Landslide Proneness of Kerala, India A Geospatial study." Landslides **18**(4): 1245-1258.

Reichenbach, P., M. Rossi, B. D. Malamud, M. Mihir and F. Guzzetti (2018). "A review of statistically-based landslide susceptibility models." Earth-Science Reviews **180**: 60-91.

Richhariya, B., M. Tanveer and A. H. Rashid (2020). "Diagnosis of Alzheimer's disease using universum support vector machine based recursive feature elimination (USVM-RFE)." Biomedical Signal Processing and Control **59**: 101903.

Romero, A., C. Gatta and G. Camps-Valls (2015). "Unsupervised deep feature extraction for remote sensing image classification." IEEE Transactions on Geoscience and Remote Sensing **54**(3): 1349-1362.

Rong, G., K. Li, L. Han, S. Alu, J. Zhang and Y. Zhang (2020). "Hazard Mapping of the Rainfall–Landslides Disaster Chain Based on GeoDetector and Bayesian Network Models in Shuicheng County, China." Water **12**(9): 2572.

Sajinkumar, K., S. Anbazhagan, A. Pradeepkumar and V. Rani (2011). "Weathering and landslide occurrences in parts of Western Ghats, Kerala." Journal of the Geological Society of India **78**(3): 249-257.

Sameen, M. I., B. Pradhan and S. Lee (2020). "Application of convolutional neural networks featuring Bayesian optimization for landslide susceptibility assessment." CATENA **186**: 104249.

Senouci, R., N.-E. Taibi, A. C. Teodoro, L. Duarte, H. Mansour and R. Yahia Meddah (2021). "GIS-based expert knowledge for landslide susceptibility mapping (LSM): case of mostaganem coast district, west of Algeria." Sustainability **13**(2): 630.

Shaheen, H., S. Agarwal and P. Ranjan (2020). MinMaxScaler binary PSO for feature selection. First international conference on sustainable technologies for computational intelligence, Springer.

Song, J., Y. Wang, Z. Fang, L. Peng and H. Hong (2020). "Potential of ensemble learning to improve tree-based classifiers for landslide susceptibility mapping." IEEE Journal of Selected Topics in Applied Earth Observations and Remote Sensing **13**: 4642-4662.

Sun, D., H. Wen, Y. Zhang and M. Xue (2021). "An optimal sample selection-based logistic regression model of slope physical resistance against rainfall-induced landslide." Natural Hazards **105**(2): 1255-1279.

Sun, D., J. Xu, H. Wen and D. Wang (2021). "Assessment of landslide susceptibility mapping based on Bayesian hyperparameter optimization: A comparison between logistic regression and random forest." Engineering Geology **281**: 105972.

Tanyu, B. F., A. Abbaspour, Y. Alimohammadlou and G. Tecuci (2021). "Landslide susceptibility analyses using Random Forest, C4.5, and C5.0 with balanced and unbalanced datasets." CATENA **203**: 105355.

Vasu, N. N. and S.-R. Lee (2016). "A hybrid feature selection algorithm integrating an extreme learning machine for landslide susceptibility modeling of Mt. Woomyeon, South Korea." Geomorphology **263**: 50-70.

Vishnu, C., K. Sajinkumar, T. Oommen, R. Coffman, K. Thrivikramji, V. Rani and S. Keerthy (2019). "Satellite-based assessment of the August 2018 flood in parts of Kerala, India." Geomatics, Natural Hazards and Risk **10**(1): 758-767.

Wang, H., L. Zhang, H. Luo, J. He and R. Cheung (2021). "AI-powered landslide susceptibility assessment in Hong Kong." Engineering Geology **288**: 106103.

Wang, Y., Z. Fang and H. Hong (2019). "Comparison of convolutional neural networks for landslide susceptibility mapping in Yanshan County, China." Science of The Total Environment **666**: 975-993.

Wang, Y., K. Wang, M. Zhang, T. Gu and H. Zhang (2023). "Reliability-enhanced surrogate-assisted particle swarm optimization for feature selection and hyperparameter optimization in landslide displacement prediction." Complex & Intelligent Systems: 1-31.

Yao, X., L. Tham and F. Dai (2008). "Landslide susceptibility mapping based on support vector machine: a case study on natural slopes of Hong Kong, China." Geomorphology **101**(4): 572-582.

Zhao, P., Z. Masoumi, M. Kalantari, M. Aflaki and A. Mansourian (2022). "A GIS-Based Landslide Susceptibility Mapping and Variable Importance Analysis Using Artificial Intelligent Training-Based Methods." Remote Sensing **14**(1): 211.

Zhao, X. and W. Chen (2020). "Optimization of computational intelligence models for landslide susceptibility evaluation." Remote Sensing **12**(14): 2180.

Zhao, Z., J. Chen, K. Xu, H. Xie, X. Gan and H. Xu (2021). "A spatial case-based reasoning method for regional landslide risk assessment." International Journal of Applied Earth Observation and Geoinformation **102**: 102381.

Zhou, J., Y. Qiu, D. J. Armaghani, W. Zhang, C. Li, S. Zhu and R. Tarinejad (2021). "Predicting TBM penetration rate in hard rock condition: A comparative study among six XGB-based metaheuristic techniques." Geoscience Frontiers **12**(3): 101091.

Zhou, X., H. Wen, Y. Zhang, J. Xu and W. Zhang (2021). "Landslide susceptibility mapping using hybrid random forest with GeoDetector and RFE for factor optimization." Geoscience Frontiers **12**(5): 101211.